\documentclass[11pt]{article}

\usepackage[preprint]{acl}

\usepackage{times}
\usepackage{latexsym}

\usepackage{float}
\usepackage{stfloats}

\usepackage[T1]{fontenc}

\usepackage[utf8]{inputenc}

\usepackage{microtype}
\usepackage{booktabs}

\usepackage{inconsolata}

\usepackage{graphicx}

\usepackage{booktabs}
\usepackage{multirow}
\usepackage{tabularx}
\usepackage{hyperref}

\title{Joint Detection of Constructive Speech, Hate Speech and Agreement in Online Conversations}

\author{Ida Dencker \\
   Aarhus University \\
  \texttt{ida3mail@gmail.com} \\\And
  Simon Karg \\
  Aarhus University \\
  \texttt{simonkarg@ps.au.dk} \\\And
  Kenneth Enevoldsen \\
  Aarhus University \\
  \texttt{kenneth.enevoldsen@cas.au.dk
} \\
  }

\begin{document}

\maketitle
\begin{abstract}
Online hate speech is an ever-growing issue with negative impacts on both groups and individuals, potentially causing users to withdraw from online conversations and debate. With some prominent social media platforms reducing their moderation efforts, constructive speech has emerged as an alternative strategy for countering hate. Automated detection of constructive and hateful speech can aid monitoring and improve understanding of their prevalence and online dynamics. In this work, we collect and annotate a large dataset of English and Danish social media comments for hateful tone, constructive tone, and agreement with parent comments. We fine-tune several multilingual LLMs to classify along these three dimensions and compare context-enriched and non-context-enriched model architectures. The E5-large LLM performs best on both the context-enriched (82\% average F1) and non-context-enriched (81.6\% average F1) classification tasks, demonstrating the potential of multilingual LLMs for scalable detection of hateful and constructive speech across languages.
\end{abstract}

\maketitle

\begin{figure}[t]
    \centering
    \includegraphics[width=.9\linewidth]{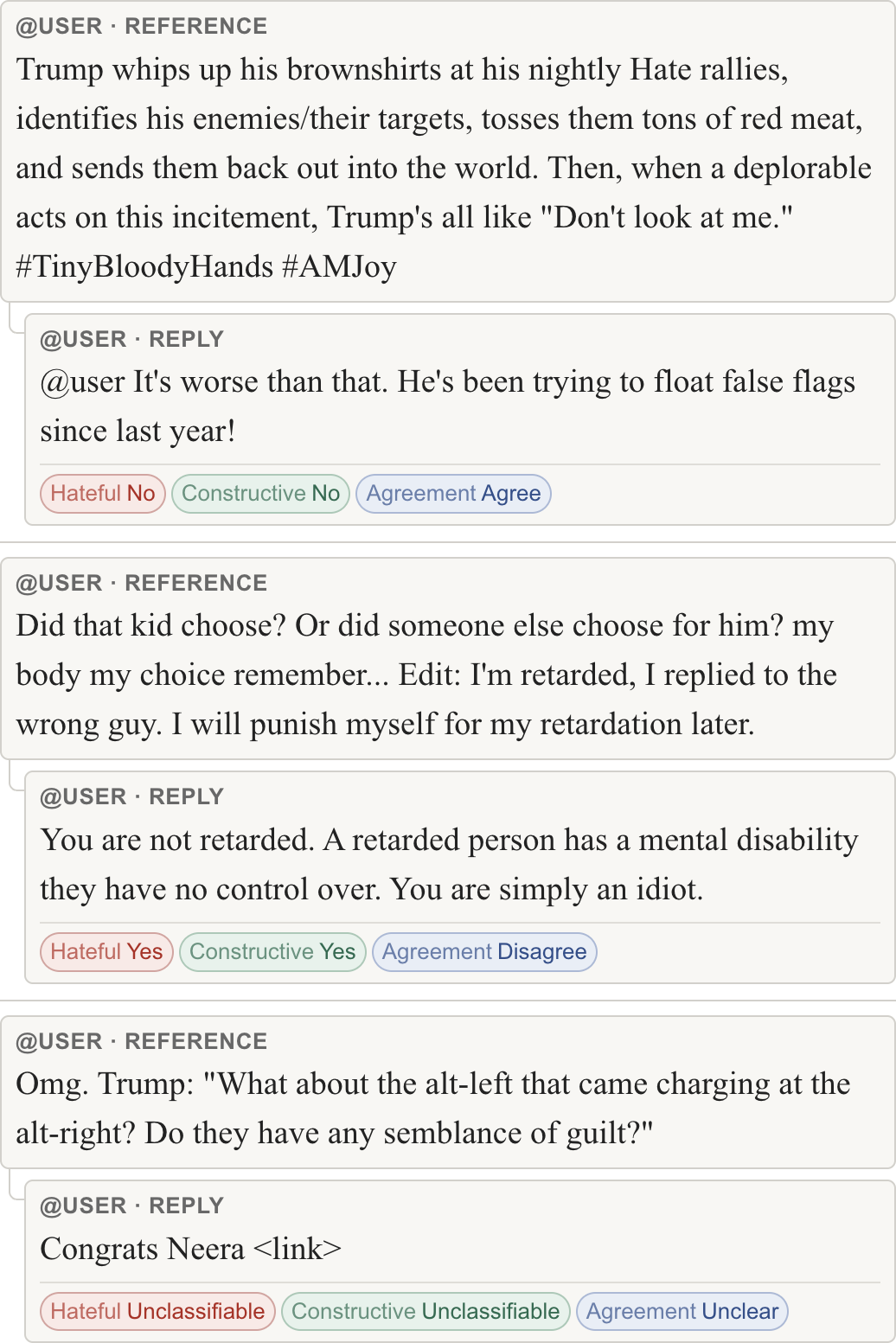}
    \caption{Example conversations with annotations.}
    \label{fig:data-structure}
\end{figure}

\section{Introduction}
Online spaces are an important part of everyday life in the 21st century. Approximately two-thirds of the world's population uses social media, and this number continues to grow \citep{kemp2022digital}. Though online conversations can serve as healthy forums for sharing, expressing, interacting, and debating across countries and cultures, they can also foster hate speech harming individuals and groups. A UNESCO \citeyearpar{unesco2025worldtrends} report reveals that 67\% of internet users have encountered hate speech online. 
In 2023, nearly half (49\%) of young active internet users aged 16–29 in EU countries reported encountering hostile or degrading messages online, with Denmark recording the second-highest share at 68\%. Furthermore, among this age group, “political or social views” was the most frequently cited reason for why individuals or groups were targeted, accounting for 35\% across the EU. This category was particularly prominent in Estonia (60\%), followed by Finland (56\%) and Denmark (49\%) \citep{eurostat2024hostile}.

Exposure to online hate and violent content has several negative personal and societal consequences. At the individual level, it is associated with lower well-being \citep{oksanen2014exposure} and happiness \citep{keipi2018exposure}, as well as increased aggression \citep{anderson2001effects} and reduced empathy \citep{funk2004violence}. It can also lead to stress \citep{saha2019prevalence}, distress  \citep{leets2002experiencing}, sadness \citep{reichelmann2021hate,keighley2022hate}, and depressive symptoms \citep{madriaza2025exposure}. At the societal level, online hate speech discourages participation in public discourse. As more and more individuals withdraw from debates \citep{marzkeda2025diversity,gelber2016evidencing,chen2017online}, distance themselves from online spaces \citep{pabian2026sa}, or engage in self-censorship \citep{celuch2024self}, the democratic potential of online spaces is threatened, as these forums become more and more polarized \citep{RN3545}. Such concerns are also reflected in current policy debates about particularly young people's use of social media. For instance, Australia has in 2024 introduced a minimum age requirement for social media use \citep{australia2024socialmediaage}, and similar discussions have emerged in Denmark and the EU more generally.

While most users scroll by or withdraw from hateful content \citep{RN3350}, others choose to act against it in various ways \citep{RN3487}. One of the most promising bystander reactions for reducing the negative impact of online hate is counterspeech, i.e., written responses to aiming to counter hateful content. Counterspeech can take different forms, with research suggesting varying effectiveness at reducing future hostility \citep{hangartner2021empathy, RN3576,Lasser2025CollectiveModeration}. In this work, we distinguish between constructive and non-constructive forms of counterspeech, with constructive counterspeech defined as responding in a generally prosocial way. In particular, constructive counterspeech seeks to challenge hate by signaling its disapproval, and/or siding with, shielding, or comforting an individual victim or victimized group. This can be achieved in various ways, such as trying to evoke empathy, calling out hateful behavior, correcting misperceptions, or generally calling for a better culture of debate \citep{RN3487}. Research has shown that constructive counterspeech is more effective than non-constructive (e.g., hateful or ridiculing) forms of counterspeech \citep{albanyan2023NotallCounterhate,Alyahya2025HatredIgnorance}.

Understanding (constructive) counterspeech is particularly important in a time of growing platform reluctance toward moderation practices. In early 2025, Meta announced a shift toward greater emphasis on free expression and reduced content moderation \citep{kaplan2025morespeech}, adopting an approach similar to that of X \citep{x2024defendingrights}. In contrast, a large share of internet users believe that hate speech should be addressed by social media platforms themselves \citep{ipsosunesco2023disinformation}. In this context, crowd-sourced moderation through counterspeech may become increasingly important.

To understand these online dynamics, it is necessary to monitor both hate speech and various forms of counterspeech at scale. Natural language processing (NLP) methods offer a promising approach for automatically detecting such content, enabling researchers to examine the prevalence of hate and constructive speech, as well as whether constructive speech improves online discussions. To enable such analysis, we introduce a multilingual, multidimensional classifier that jointly labels texts for hate speech, constructive speech, and agreement with parent comments.

\subsection{Our Contribution}
In this work, we collect and annotate a large dataset of English (a high-resource language) and Danish (a mid-to-low-resource language) \cite{joshi2020lingdiversity} social media comments along three dimensions: hateful (yes/no/unclassifiable), constructive speech (yes/no/unclassifiable), and agreement (agree/disagree/unclear). We fine-tune four multilingual LLMs for the two detection setups (context-enriched and non-context-enriched), and conduct a comparative evaluation across models.

This is, to the best of our knowledge, the first study to jointly model hate speech, constructive speech, and agreement in a multidimensional and non-exclusive framework, allowing instances to, simultaneously express hate and constructive speech. However, a similar idea has been explored by \citet{albanyan2022Pinpointing}, albeit with a more narrowly focused approach.

Together, the three dimensions allow the tracking of the precise dynamics of hate and (constructive) counterspeech as they unfold in online comment sections, using just a single classification model. Furthermore, this work introduces the first annotated dataset and constructive speech classification model in Danish, addressing a key gap in research on mid-to-low-resource languages.

\section{Previous Work}

\subsection{Challenges in automated detection}
Automatically detecting hate speech remains a challenging task. Hate speech is often ambiguous and highly dependent on context, making it difficult to distinguish between genuinely hateful content and harmless uses of similar language \citep{wilson2020hate,markov2022role,sheth2022toxicity}. In addition, hateful expressions continuously evolve through new slang, coded language, and platform-specific conventions \cite{perez2023assessing}. As a result, effective detection cannot rely solely on the presence of profanity or offensive keywords (explicit hate), but should ideally also detect more implicit or subtle forms of hate \cite{gao2017detecting}. Detection systems used in, for example, moderation must also balance the identification of harmful content with the protection of legitimate expression, reflecting broader tensions between content regulation and free speech.

Detecting constructive speech and other forms of pro-social speech presents similar challenges. Definitions of constructive speech are inherently subjective, and responses perceived as constructive by some users may be viewed as condescending, hostile, or ineffective by others \cite{ping2025perceiving}. 
Furthermore, interpretations of constructive communication vary across cultural and social contexts, complicating the development of consistent annotation schemes and automated detection systems \cite{fesaghandis2026multilingual,fesaghandis2026parscn}.

We note that, in this work, we focus on encoder-based classification models. While the field of hate speech detection has seen rapid development accompanying the ever-growing capabilities of LLMs (e.g., using decoder models as single or few-shot classifiers \citep{rathje2024GPTiseffective}), the effectiveness of these approaches when compared to more classical approaches of using fine-tuned encoder models remains an open question \citep{zhang-etal-2025-bert-like}, and come with their own set of issues \citep{fasching-lelkes-2025-model}.  We believe that for the foreseeable future, there remains a need for specialized lightweight models that can be run locally on smaller machines to enable the analysis of large-scale data at a reasonable speed and cost. However, we benchmark our own model against Jev \citep{typesafe2026jev}, a so-called "decision model" optimized to make fast text classifications. This benchmark test showed that our own fine-tuned models substantially outperformed Jev, particularly in Danish (see Appendix).

\subsection{Detecting Hate Speech}
Detecting hate speech has been widely approached as a supervised classification problem, with methods evolving from traditional machine learning to neural networks and more recently transformer-based models \citep{alkomah2022literature}. A large body of hate speech detection research has focused on English 
\citep[e.g.,][]{burnap2016us,kshirsagar2018predictive,malmasi2017detecting}. 

Non-English hate speech detection has also been explored, although to a more limited extent and often in language-specific settings
\citep[e.g.,][]{del2017hate,pereira2019detecting}. 

When exploring multilingual hate speech detection, one common approach has been to mix English with Non-English language(s) \citep[e.g.,][]{ahmad2025ua,farooqi2021leveraging,mnassri2024multilingual,siddiqui2024fine}. 

Hate speech detection has also been explored in Nordic languages, including Norwegian \citep{hashmi2024multi}, Swedish \citep{fernquist2019study}, and Danish \citep{sigurbergsson2020offensive}, although research in these languages remains more limited.

\subsection{Detecting Constructive Speech}
Detecting constructive speech remains a relatively underdeveloped NLP task, largely due to the absence of a shared definition or taxonomy and the limited availability of high-quality annotated (multilingual) datasets 
\citep{fesaghandis2026multilingual, mathew2019thou}. 

Existing work has primarily focused on English data, where early studies demonstrated that constructive speech can be detected with moderate performance using supervised models \citep{mathew2018analyzing,he2021racism,jacobs2022automatic} . Subsequent research has highlighted the importance of conversational context, with studies such by~\citet{yu2022hate},~\citet{fialho2024counter}, or \citet{albanyan2022Pinpointing} who incorporated preceding discourse in their classification setup (which we call context-enriched in our approach), leading to improved performance. 

Taking a different angle, some research focusses on differentiating counterspeakers from hate speakers. That is, rather than classifying a single piece of text, these studies take individual users as the focus of analysis \citep{garland2020countering,mathew2020InteractionDynamicsUsers}. 

As perhaps the most ambitious attempt thus far in counterspeech detection,  \citet{Lasser2025CollectiveModeration} trained a German classifier for counterspeech that also covers various more fine-grained counterspeech strategies (e.g., sarcasm, providing facts, offering opinions, etc.). However, since most strategies were too rare to be reliably detected, the final classifier collapses most categories into broader categories: opinion, prosocial, sarcasm, insult, or other/uninterpretable. This also echoes earlier findings by \citet{yu-etal-2023-finegrainedTaxonomy}, who reported only moderate performance in classifying counterspeech strategies.

However, most studies rely on data from a single social media platform, limiting the generalizability of findings across platforms. In addition, multilingual work remains limited and often focuses on a mix of small sets of high-resource languages \citep{chung2019conan,chung2021multilingual,Möhle2023JustCollect}.

\subsection{Limitations of Current Approaches}
Most existing systems for detecting constructive speech, toxicity, or hate speech operate on isolated comments without sufficient context, often classifying a single message without access to the surrounding conversation, such as the parent comment or the full thread. While a few recent studies incorporate conversational context \cite{yu2022hate,fialho2024counter,albanyan2022Pinpointing}, this is still not the norm. This limitation is important because the meaning of a message often depends heavily on its context and external knowledge \cite{sheth2022toxicity}. 

Furthermore, there is currently a lack of comprehensive models that capture the full dynamics of online discourse, as most existing work focuses on binary or single-label classification. In particular, hate speech and constructive speech are often studied separately, despite representing related aspects of interaction within the same conversational context. Jointly modelling these dimensions may improve computational efficiency and classification robustness by allowing the model to leverage shared information across related tasks \cite{collobert2008unified}, while also enabling more comprehensive large-scale analyses of online discourse.

Closest to our work is \citet{albanyan2022Pinpointing}, who classify replies to hateful tweets hierarchically: first as agreeing or disagreeing with the hateful parent, then, if disagreeing, whether they justify the disagreement or attack the author, and, if agreeing, whether they add further hate. Our approach differs in three ways. First, our labels apply to every comment in a conversation tree, including the source post, whereas their scheme labels only replies to hate and cannot detect initial hateful comments. Second, every dimension includes an unclassifiable label, so spam and incomprehensible comments are handled explicitly rather than forced into a yes/no decision \citep[see also][]{Lasser2025CollectiveModeration}. Finally, their model has not been released and, given X's restricted API, cannot easily be retrained.

In sum, we present a dataset and model that jointly capture multiple aspects of online interaction, including hateful, constructive, and agreement-related content. This allows for a nuanced and non-exclusive characterization of user behavior within social media conversations, while supporting context-aware and scalable analysis across languages and platforms.

\section{Methodology }

\subsection{Data Collection}
The dataset consists of pairs of texts (social media comments). The first text (called the reference text) can be any kind of text, such as a top-level post, a reply, or a quote tweet. The second text (the reply text) is always a reply to the reference text. For the non-context-enriched model, only the reply text is used as input. For the context-enriched model, both the reference text and the reply text, combined with a [SEP] token, are used as input.

\begin{figure}
    \centering
    \includegraphics[width=1\linewidth]{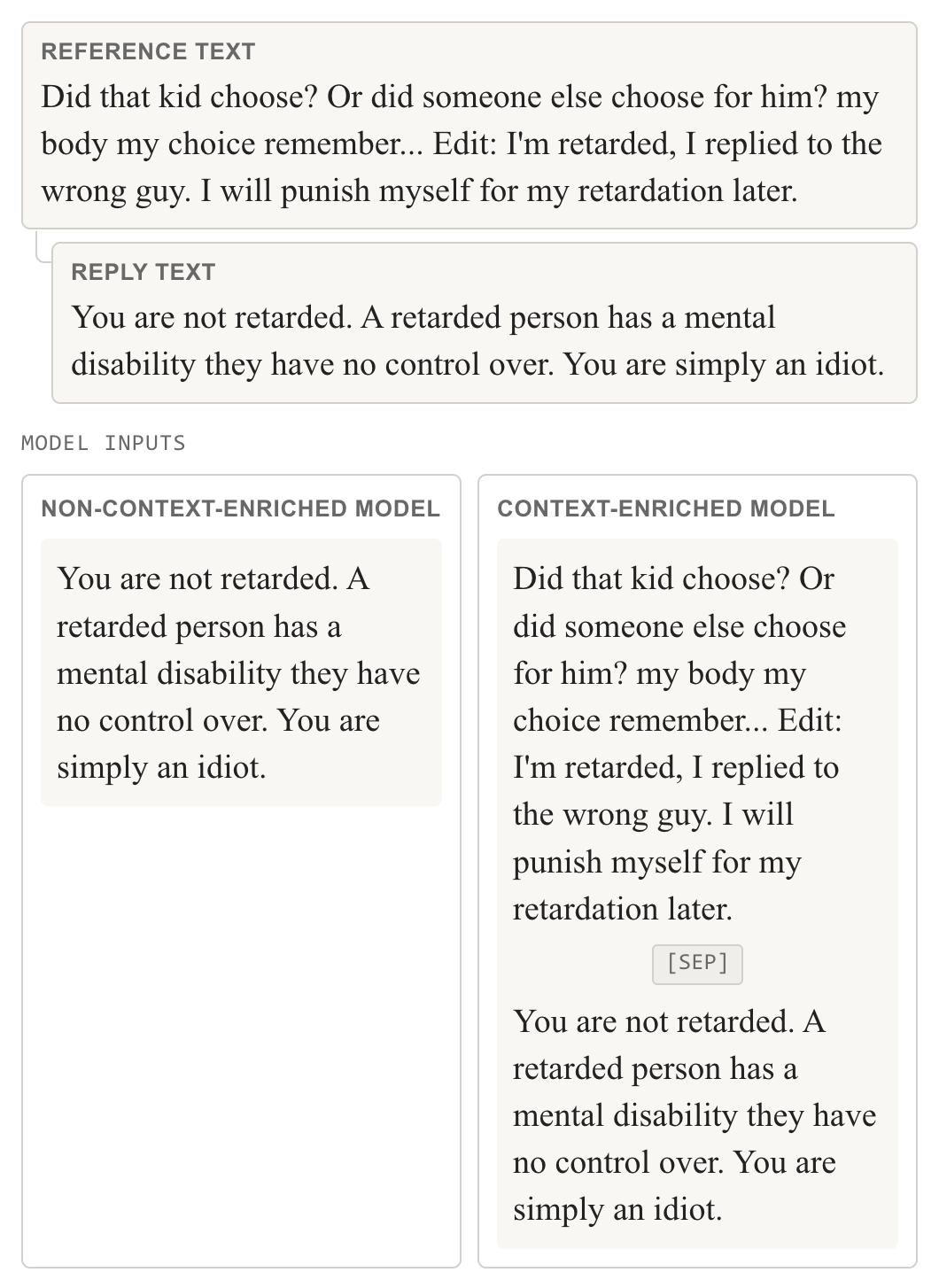}
    \caption{Example of training input for context-enriched vs non-context-enriched models.}
    \label{fig:input}
\end{figure}

We aimed to include texts in multiple languages, sourced from different platforms and covering a wide range of topics to create a diverse and appropriate dataset for annotation. Data was collected from Twitter (now X), Reddit, and Gab\footnote{Gab is a far-right social media platform founded in 2016 that positions itself as a free-speech alternative to mainstream networks, known for minimal content moderation and for a user base that includes many extremist communities.}. 

The total dataset consists of 7,388 pairs, of which 6,199 (84\%) are English and 1189 (16\%) are Danish. The pairs were collected and annotated in several stages. For the English texts, we sampled 5839 text pairs from a larger corpus of tweets about US politics \cite{rasmussen2023eventdriven} (after annotation, 5199 pairs remained). A further 500 text pairs each from Reddit and Gab were taken from a collection by~\citet{qian2019intervene}. Different strategies were employed for collecting high-quality and relevant source data. While the ~\citet{rasmussen2023eventdriven} data were based on Twitter data from 2839 users whom they later surveyed, \citet{qian2019intervene} used hateful keywords to identify relevant conversations. The English texts were posted between 2008 and 2019, with most tweets dating from 2016 to 2019 and revolving mainly around American politics and Trump’s presidency. 

The 1189 Danish texts were sampled from a larger collection of Danish tweets \citep{Rasmussen2024DanishDataOfflineRoots} collected using the official Twitter API and the R package academictwitteR \cite{barrie2021academ}. The dataset primarily consists of tweets posted by Danish individuals (i.e., excluding official and celebrity accounts). The Danish tweets were sampled and labelled in two stages (see data annotation below), with tweets dating from 2009 to 2022, though most from 2020 to 2022. This latter period encompassed several significant events, including the 2022 Danish general election. Most Danish tweets are likewise about politics; however, other casual topics also occur in the data.
A dataset overview is provided in Table \ref{tab:dataset}.


\begin{table*}[t]
\centering
\small
\begin{tabularx}{\textwidth}{@{}l l r l >{\raggedright\arraybackslash}X >{\raggedright\arraybackslash}X@{}}
\toprule
\textbf{Source} & \textbf{Language} & \textbf{Size} & \textbf{Timeframe} & \textbf{Collection strategy} & \textbf{Thematics} \\
\midrule
Twitter \citep{rasmussen2023eventdriven} & English & 5200 & 2008--2019 & Data from 2839 surveyed Twitter users & COVID-19, Donald Trump, American politics \\
Reddit \citep{qian2019intervene}   & English & 500  & 2008--2019 & Identify hateful subreddits and hate keywords & Diverse but hateful \\
Gab \citep{qian2019intervene}      & English & 500  & 2008--2019 & Hate keywords & Diverse but hateful \\
Twitter \citep{Rasmussen2024DanishDataOfflineRoots} & Danish & 1189 & 2009--2022 & Data from 4931 Danish Twitter users & Politics and everyday life \\
\midrule
Combined dataset & English \& Danish & 7388 & 2008--2022 & & \\
\bottomrule
\end{tabularx}
\caption{Dataset overview.}
\label{tab:dataset}
\end{table*}

\subsection{Data Annotation}
English and Danish data were annotated separately following the same codebook, available as supplementary material.\footnote{Codebooks were identical except that the English version also included a separate labelling dimension ("tweet type") that we dropped for Danish, as it proved unnecessary for this study} The English annotation setup used Qualtrics software (Provo, UT), while Danish annotation was carried out using a Shiny app \citep{shiny2025}.

The English dataset was coded by a team of five coders, with an initial training period covering the first 200 tweets. During this period, the five coders worked together with a broader team of researchers to iteratively improve the codebook and to discuss and resolve disagreements. The five coders were Danish political science students aged 20-25, all highly proficient in English (3 female, 2 male). The Danish dataset was coded by two separate teams. The initial batch of 189 tweets was coded by a research team at the Department of Political Science of Aarhus University as a proof of concept. The second batch of Danish tweets was coded by three Danish university students aged 20-24 (all female). These three coders were initially trained on a set of previously coded English tweets, and then further trained on a set of 150 Danish tweets, after which they met with the research team to discuss disagreements. All student coders were paid the standard rate for student assistants at Danish universities (approx. \$18 per hour).

The coders were instructed to determine whether each text contained hateful (yes, no, unclassifiable) and/or constructive (yes, no, unclassifiable) elements, as well as to classify agreement (agree, disagree, unclear)  relative to its reference (parent) text. 


\begin{table*}[t]
\centering
\small
\begin{tabular}{@{}l rrr rrr rrr@{}}
\toprule
\multirow{2}{*}{\textbf{Language}} & \multicolumn{3}{c}{\textbf{Constructive}} & \multicolumn{3}{c}{\textbf{Hate}} & \multicolumn{3}{c}{\textbf{Agreement}} \\
\cmidrule(lr){2-4} \cmidrule(lr){5-7} \cmidrule(l){8-10}
 & Y & N & U & Y & N & U & A & D & U \\
\midrule
Danish  & 13.5\% & 81.7\% & 4.8\% & 10.5\% & 84.7\% & 4.8\% & 25.8\% & 38.7\% & 35.5\% \\
English & 20.7\% & 72.5\% & 6.9\% & 16.3\% & 77.0\% & 6.7\% & 38.8\% & 37.9\% & 23.3\% \\
\midrule
Total & 19.5\% & 74.0\% & 6.5\% & 15.3\% & 78.2\% & 6.4\% & 36.7\% & 38.0\% & 25.3\% \\
\bottomrule
\end{tabular}
\caption{Class distribution across languages and annotation dimensions. Y = Yes, N = No, U = Unclassifiable for constructive speech and hateful; A = Agree, D = Disagree, U = Unclear for agreement.}
\label{tab:classimbalance}
\end{table*}

Hateful elements are defined as any expression of a motivation to degrade, harass, or intimidate individuals or collectives based on political differences or discriminatory views. In practice, annotators relied on a set of guiding principles: (i) demeaning or dehumanizing language directed at individuals or groups (e.g., insults or derogatory labels) is typically considered hateful; (ii) explicit statements of hate toward a person or group are labeled as hateful, unless clearly used in a non-literal or ironic manner; (iii) the use of slurs or highly uncivil language is often indicative of hate; and (iv) merely accusing someone of being hateful does not in itself constitute hate unless accompanied by other degrading elements.

Constructive elements imply a prosocial attitude, signaling disapproval of hateful or hostile behavior, or aiming to side with, shield, or comfort an individual victim or victimized group. These include responses that aim to promote more respectful discourse or provide reasoned criticism. However, unsupported accusations (e.g., labeling someone as racist without justification) are not considered constructive, whereas reasoned arguments explaining such claims are.

Importantly, these two categories are not mutually exclusive, meaning that a single text can simultaneously contain both hateful and constructive elements. The texts that could not be meaningfully classified as containing either hateful or constructive elements were labeled unclassifiable.

In addition, annotators labeled agreement (agree, disagree, unclear) with respect to the reference text. This label captures whether the reply aligns with the stance of the reference. Agreement is interpreted at the level of intent rather than surface form (e.g., ironic agreement is coded as disagreement). Unlike hate and constructiveness, which can co-occur, ,agreement labels are mutually exclusive, and unclear or mixed cases are labeled “unclear”.

For the English Twitter data containing 5,839 pairs, each text pair was coded by at least three coders. 
Inter-rater agreement for the different dimensions was relatively low (Krippendorff’s alpha: hateful: 0.46; constructive: 0.45; agreement: 0.50), reflecting the well-documented difficulty of the task \citep{fonseca2024analyzing,zhang2023taxonomy}. Disagreements in assigned labels were resolved by majority vote, with examples showing three-way ties excluded (resulting in a final sample of 5,199 tweets). The 1,000 Reddit and Gab data was later coded by a single coder from the initial team of five coders. This was done because this data was pre-selected to be prosocial responses to hate speech by \citet{qian2019intervene}, meaning that the coding mainly served to check whether the definitions used in this current study aligned with the conceptualizations of hateful and constructive counterspeech employed by \citet{qian2019intervene}.

For the larger Danish dataset containing 1,000 pairs, three trained coders annotated the data, such that each text pair was coded by all three coders. Inter-rater agreement for the different dimensions was broadly similar to that for the English data (Krippendorff’s alpha: hateful: 0.58; constructive: 0.36; agreement: 0.55). Disagreements were first jointly assessed, with remaining undecided cases resolved by the data collection leader. The smaller Danish dataset containing 189 pairs was labelled by four researchers from the research group, with ties being jointly resolved.

Class imbalance was identified across two dimensions (hateful and constructive). The imbalance is present both within each language subset and in the combined dataset. The majority class dominates in all cases. For example, non-hateful content constitutes approximately 77–85\% of the data across languages, while hateful instances remain relatively scarce (10–16\%). Similarly, non-constructive speech instances account for roughly 72–82\%. Such class imbalance introduces a risk that models will disproportionately learn to predict the majority class, potentially achieving high overall accuracy while performing poorly on minority classes. This is particularly critical in tasks such as hate detection, where the minority class is often of greatest interest.
An overview of class imbalance is provided in Table \ref{tab:classimbalance}.

The observed class imbalance is consistent with previous research showing that hateful and toxic content constitutes only a minority of online communication \cite{gennaro2025hatedistribution,kantor2020progress}.
At the same time, we believe our dataset contains a sufficient number of instances for the minority classes to enable meaningful model training and evaluation, balancing representativeness with practical usability.

\subsection{Classifier base models}
We conducted experiments using four pretrained multilingual encoder-based language models (all openly available via Hugging Face):

\begin{enumerate}
    \item XLM-RoBERTa-large
    \cite{conneau2020xlm}
    \item multilingual-e5-large 
    \cite{wang2024multilinguale5}
    \item Twitter-XLM-RoBERTa-base 
    \cite{barbieri2022xlmt}
    \item mmBERT-base 
    \cite{marone2025mmbert}
\end{enumerate}

To support reproducibility, we supply the exact Hugging Face IDs and revisions in the appendix.
These models were selected for their strong multilingual capabilities and prior evidence of effective cross-lingual transfer without explicit language-specific training, enabling efficient knowledge transfer
\citep{barbieri2022xlmt,conneau2020xlm,marone2025mmbert,wang2024multilinguale5}.

We prioritized models that can be deployed locally with relatively low computational overhead, supporting practical applicability in real-world settings. For each base model, we fine-tune two variants: one using only the reply text, and one incorporating contextual information in the form of the reference (parent) text. This yields eight model configurations in total. 


\begin{table*}[t]
\centering
\resizebox{1\textwidth}{!}{
\begin{tabular}{@{}l rrrr rrrr rrrr r@{}}
\toprule
\multirow{2}{*}{Model} & \multicolumn{4}{c}{Constructive} & \multicolumn{4}{c}{Hate} & \multicolumn{4}{c}{Agree} & \multirow{2}{*}{Avg. F1} \\
\cmidrule(lr){2-5} \cmidrule(lr){6-9} \cmidrule(lr){10-13}
 & Y & N & U & F1 & Y & N & U & F1 & A & D & U & F1 & \\
\midrule
\multicolumn{14}{@{}l}{\textit{Non-context-enriched}} \\
xlm-roberta-large         & .79 & .92 & .67 & \textbf{.88} & .63 & .91 & .69 & \textbf{.86} & .72 & .79 & .52 & .70 & .813 \\
multilingual-e5-large     & .77 & .92 & .77 & \textbf{.88} & .65 & .91 & .68 & \textbf{.86} & .70 & .80 & .56 & \textbf{.71} & \textbf{.816} \\
twitter-xlm-roberta-base  & .76 & .92 & .65 & \textbf{.88} & .58 & .91 & .68 & .85 & .70 & .79 & .50 & .69 & .806 \\
mmBERT-base               & .74 & .91 & .58 & .86 & .55 & .91 & .59 & .85 & .65 & .76 & .52 & .66 & .790 \\
\midrule
\multicolumn{14}{@{}l}{\textit{Context-enriched}} \\
xlm-roberta-large         & .76 & .92 & .67 & .87 & .62 & .91 & .69 & \textbf{.86} & .73 & .81 & .55 & \textbf{.72} & .816 \\
multilingual-e5-large     & .79 & .93 & .60 & \textbf{.88} & .67 & .91 & .62 & \textbf{.86} & .74 & .81 & .54 & \textbf{.72} & \textbf{.820} \\
twitter-xlm-roberta-base  & .72 & .91 & .59 & .86 & .53 & .90 & .60 & .83 & .67 & .76 & .53 & .67 & .786 \\
mmBERT-base               & .73 & .91 & .55 & .87 & .37 & .90 & .57 & .83 & .67 & .77 & .55 & .68 & .793 \\
\bottomrule
\end{tabular}
}
\caption{Model performance across constructive speech, hate speech, and agreement classification, evaluated with and without contextual enrichment. F1 scores are reported for each classification task alongside the average F1 score across tasks. Bold values indicate the best score in each category, with ties highlighted. For constructive speech and hate speech, Y = Yes, N = No, and U = Unclassifiable. For agreement, A = Agree, D = Disagree, and U = Unclear.}
\label{tab:results}
\end{table*}

\subsection{Task setup}

The problem is formulated as three independent multi-class classification tasks for the three dimensions (hateful, constructive, agreement), each with three mutually exclusive labels (see data annotation section above).

The model produces a vector of nine logits for each input, which are partitioned into three groups of three, one for each task, and then converted into probabilities. At the prediction stage, the label with the highest probability within each group of logits is taken as the predicted label. 

Model selection is based on the highest average validation accuracy across the three tasks. All base models are fine-tuned on the collected dataset and optimized through hyperparameter tuning.

\subsection{Hyperparameters}
Models were trained using the same 80/20 training–validation split. Training was performed with a batch size of 16 for up to 20 epochs, with early stopping applied based on average validation performance (patience = 4 epochs). Optimization was carried out using AdamW with a learning rate of $2 \times 10^{-6}$ and a weight decay of 0.1. The models were trained using a multi-task objective, computing the cross-entropy separately for each of the three classification tasks and averaging the three losses.
We used a plateau-based learning-rate decay \cite{bottou2012stochastic}, which reduced the learning rate by a factor of 0.5 after two epochs without improvement in validation loss, down to a minimum learning rate of $1 \times 10^{-7}$.

\subsection{Data preprocessing \& tokenization}
A simple preprocessing pipeline was employed. Usernames were replaced with @user, except for usernames appearing more than 100 times in the dataset (keeping public figures and prominent accounts). URLs were replaced with <link>. Emojis and hashtags were retained.
For non-context-enriched models, only the tokenized reply text was used as model input.
For context-enriched models, both the reference text and the reply text were tokenized and concatenated into a single sequence, separated by a [SEP] token, following prior work \cite{hickey2026assessing,yu2022hate}. The maximum sequence length was set to the default of 512 tokens for all models. The average length of reply texts was 32 tokens, while the average length of concatenated reference–reply pairs was 84 tokens.

\subsection{Results}
We report F1-scores for each dimension (hateful speech, constructive speech, and agreement), alongside the weighted average, in Table \ref{tab:results}. Confusion matrices for all models are provided in the Appendix. The supplementary materials also include detailed classification reports with precision, recall, and F1-scores for all classes and labels.

\section{Discussion}
Overall, the models achieved strong classification performance across all three tasks, indicating that multilingual language models can effectively detect hate speech, constructive speech, and agreement. 

Performance was lowest for the agreement task, suggesting that agreement and disagreement are more difficult to identify than hate and constructive speech. Across tasks, the unclassifiable/unclear labels were the most challenging to predict, followed by the positive ("yes") labels. Incorporating the reference text resulted in only marginal and inconsistent improvements across models and tasks.

The findings indicate that multilingual language models can be used to monitor both hate speech and constructive speech at scale. This could enable researchers to examine the prevalence of hateful and constructive content, as well as their dynamics and interaction within online discussions.

\section{Conclusion}
In this work, we collected and annotated a diverse social media dataset of reference–reply text pairs for joint detection of hate speech, constructive speech, and agreement in English and Danish. To the best of our knowledge, this is the first annotated dataset that jointly annotates all three dimensions in a non-exclusive framework and the first dataset for Danish constructive speech detection.

We fine-tuned four multilingual transformer models in both context-enriched and non-context-enriched settings. Overall, the models achieved strong classification performance, with average F1 scores ranging from 78.6\% to 82.0\%. The effect of including conversational context varied across models: three models showed only marginal improvements, whereas Twitter-XLM-RoBERTa-base showed a decrease of 2.0 percentage points. The best-performing configuration was multilingual-e5-large with context, reaching an average F1 score of 82.0\%, compared to 81.6\% without context.

These findings generally point to the potential of multilingual language models for large-scale and non-exclusive detection of hate speech, constructive speech, and agreement, including for Danish. The dataset and models presented here provide a foundation for future research and practical applications aimed at monitoring online discourse and analyzing the dynamics between hate speech and constructive responses on social media.

\section*{Limitations}
This study has three main limitations. First, the models operate exclusively on text (and rely mostly on social media comments about politics), whereas hate speech and constructive speech are often expressed through multimodal content such as images, videos, and memes \citep{chhabra2023literature, kiela2020hateful, yang2019exploring}.

Second, the annotation scheme relies on relatively broad labels (e.g., yes/no/unclassifiable), which do not distinguish between different types or strategies of hate speech and constructive speech, that have been theorized to be of varying effectiveness \citep{albanyan2022Pinpointing, albanyan2023NotallCounterhate, yu-etal-2023-finegrainedTaxonomy}. 

Third, the dataset includes only English and Danish data. While the multilingual models used in this study have demonstrated strong cross-lingual transfer capabilities, the results can be expected to be most reliable for these two languages.

Future work could address these limitations by incorporating multimodal data or utilize natively multimodal models, developing more fine-grained classification schemes, and extending or evaluating the approach to additional languages and cultural contexts.

\subsection*{Availability}
Code and supplementary material are available on GitHub\footnote{\url{https://github.com/idadencker/STANDBYs_classifier_constructive.git}}. 
The best-performing models are available via Hugging Face \footnote{\url{https://huggingface.co/AU-STANDBY/constructive_classifier_EN_DA_non_context_enriched}} \footnote{\url{https://huggingface.co/AU-STANDBY/constructive_classifier_EN_DA_context_enriched}}.
As data are derived from social media they are made available as a gated access\footnote{\url{https://huggingface.co/datasets/AU-STANDBY/annotated_constructive_dataset_DA_EN}}.

\section*{Acknowledgements}
Kenneth Enevoldsen is funded by Danish Foundation Models (4378-00001B), the European Union, Horizon Europe (101178170), the Danish National Research Foundation (DNRF193), the Aage and Johanne Louis-Hansens Foundation (25-1-17733), and the Augustinus Foundation (2025-0299).

The STANDBY research project is funded by the European Research Council (ERC) under grant agreement No. 101002251.

We thank the coders (Dagmar Vibe Huus Theilgaard , Emma Magnussen, Caroline Arild Christensen, Christian Jonasen Noer, Jesper Nielsen, Victor Nordgaard, Mette Kaad Jørgensen, Abigail Christesen, Laura Lavendal Bondesen) as well as the STANDBY research team (Lasse Lindekilde, Jesper Sommer Rasmussen, Tanja Hansen) for conceptual input as well as annotating the initial batch of Danish data, which served as a proof of concept.

We also thank Rebekah Baglini and Orla Mallon for their assistance with model development during the initial development phase.

Part of the computation done for this project was performed on the UCloud interactive HPC system, which is managed by the eScience Center at the University of Southern Denmark.

\subsection*{Author Contributions}
Ida Dencker led the model development and drafted the paper. Simon Karg carried out the data collection, designed and ran the annotation process, and conducted the literature search. Kenneth Enevoldsen provided feedback, input, and expertise throughout the project in both writing and model development.
All authors reviewed and edited the paper.

\subsection*{Declaration of Competing Interest}
The authors declare that they have no known competing financial interests or personal relationships that could have influenced the work reported in this paper.

\bibliography{custom}

\clearpage

\appendix

\section{Model IDs}
\label{sec:app:model-ids}


Table \ref{tab:model-ids} show the exact model names and IDs.

\begin{table*}[t]
\centering
\small
\renewcommand{\arraystretch}{1.15}
\begin{tabular}{@{}l l l l@{}}
\toprule
\textbf{Model} & \textbf{Hugging Face ID} & \textbf{Revision} & \textbf{Reference} \\
\midrule
XLM-RoBERTa-large & \texttt{FacebookAI/xlm-roberta-large} &
\shortstack[l]{\texttt{c23d21b0620b635a}\\\texttt{76227c604d44e43a9f0ee389}} &
\citet{conneau2020xlm} \\

multilingual-e5-large & \texttt{intfloat/multilingual-e5-large} &
\shortstack[l]{\texttt{3d7cfbdacd47fdda}\\\texttt{877c5cd8a79fbcc4f2a574f3}} &
\citet{wang2024multilinguale5} \\

Twitter-XLM-RoBERTa-base & \texttt{cardiffnlp/twitter-xlm-roberta-base} &
\shortstack[l]{\texttt{4c365f1490cb329b}\\\texttt{52150ad72f922ea467b5f4e6}} &
\citet{barbieri2022xlmt} \\

mmBERT-base & \texttt{jhu-clsp/mmBERT-base} &
\shortstack[l]{\texttt{c5955035435e2bf1}\\\texttt{21cde7f3c8863ef52ff35d82}} &
\citet{marone2025mmbert} \\
\bottomrule
\end{tabular}
\caption{Models, Hugging Face identifiers, and revisions used in the study.}
\label{tab:model-ids}
\end{table*}

\section{Codebooks}
\label{sec:app:codebook}

The codebooks in Danish and English can be found in the supplementary material. 

\clearpage
\section{Appendix: Confusion Matrices}
\label{sec:appendix}

\subsection{XLM-RoBERTa-large}

\begin{figure}[H]
    \centering

    \includegraphics[width=0.85\linewidth]{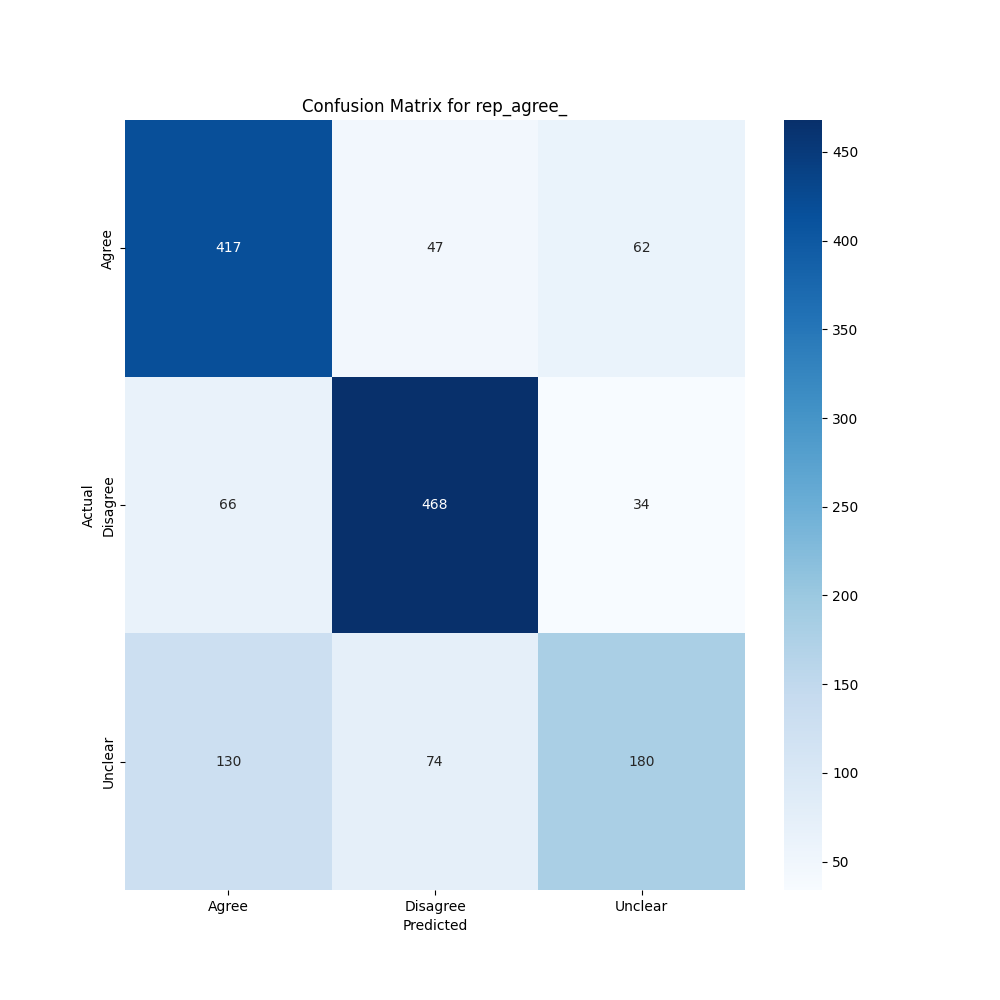}

    \vspace{0.2cm}

    \includegraphics[width=0.85\linewidth]{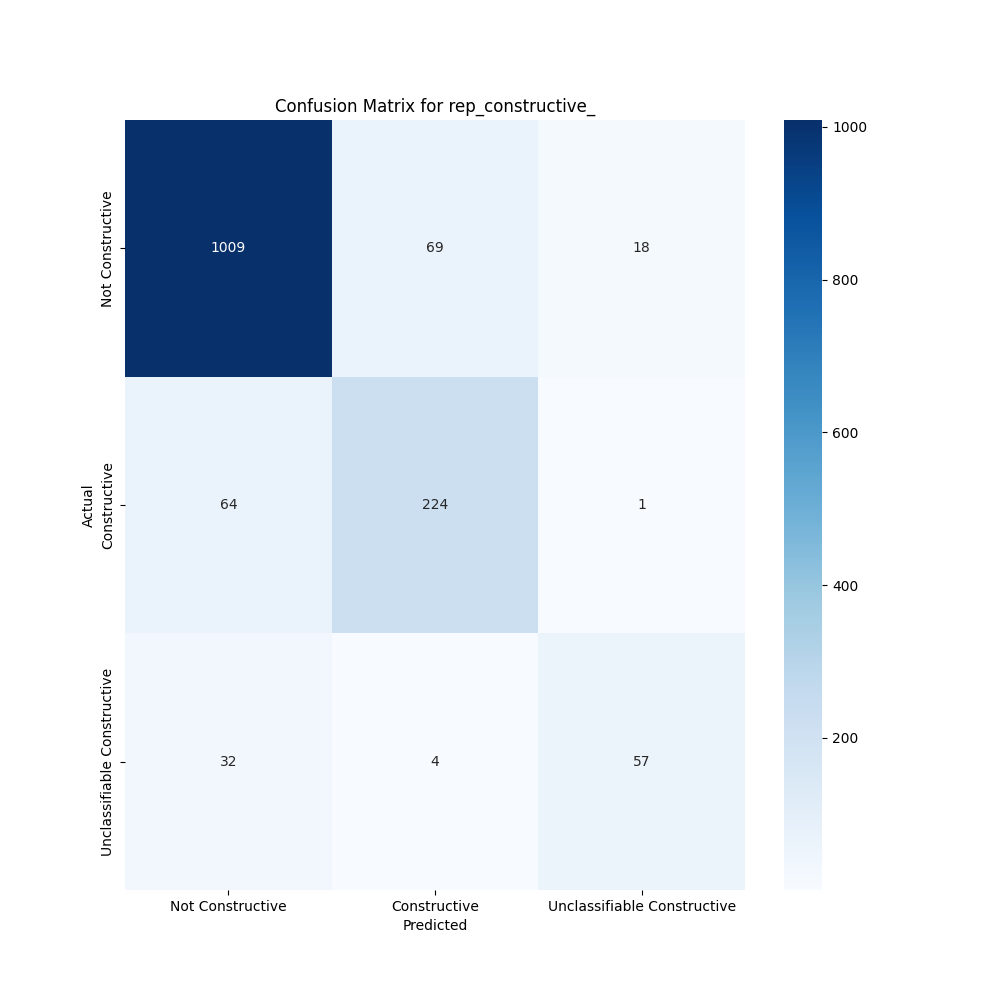}

    \vspace{0.2cm}

    \includegraphics[width=0.85\linewidth]{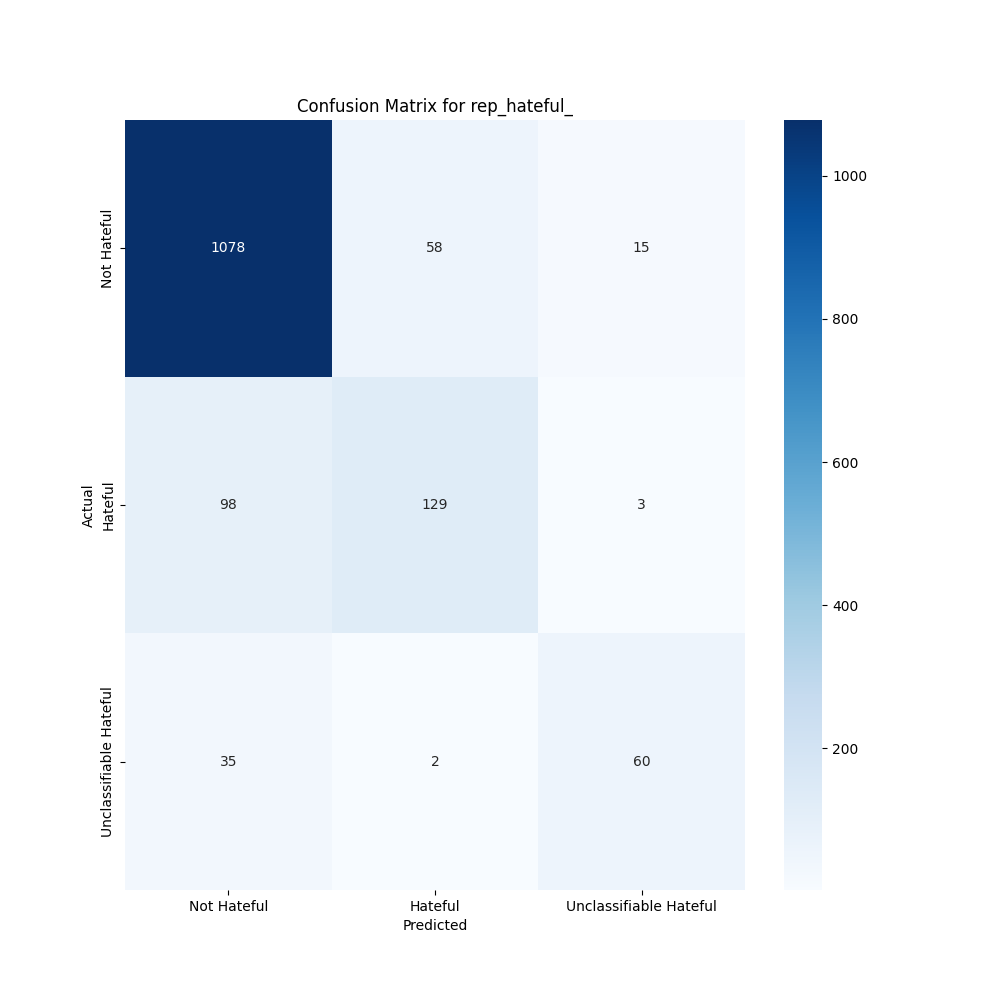}

    \caption{Confusion matrices for the context-dependent XLM-RoBERTa-large model: (a) agreement, (b) constructive speech, and (c) hateful speech.}
    \label{fig:xlm-confusion}
\end{figure}

\begin{figure}[H]
    \centering

    \includegraphics[width=0.85\linewidth]{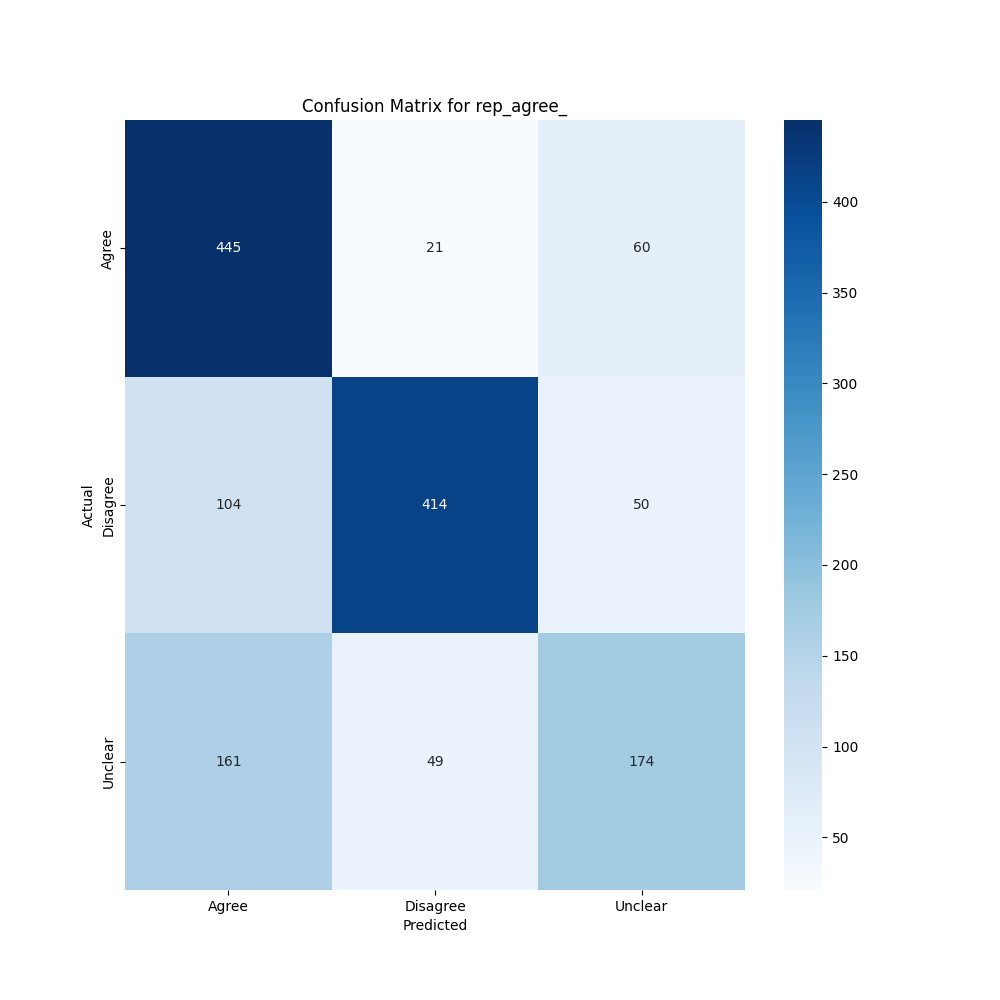}

    \vspace{0.2cm}

    \includegraphics[width=0.85\linewidth]{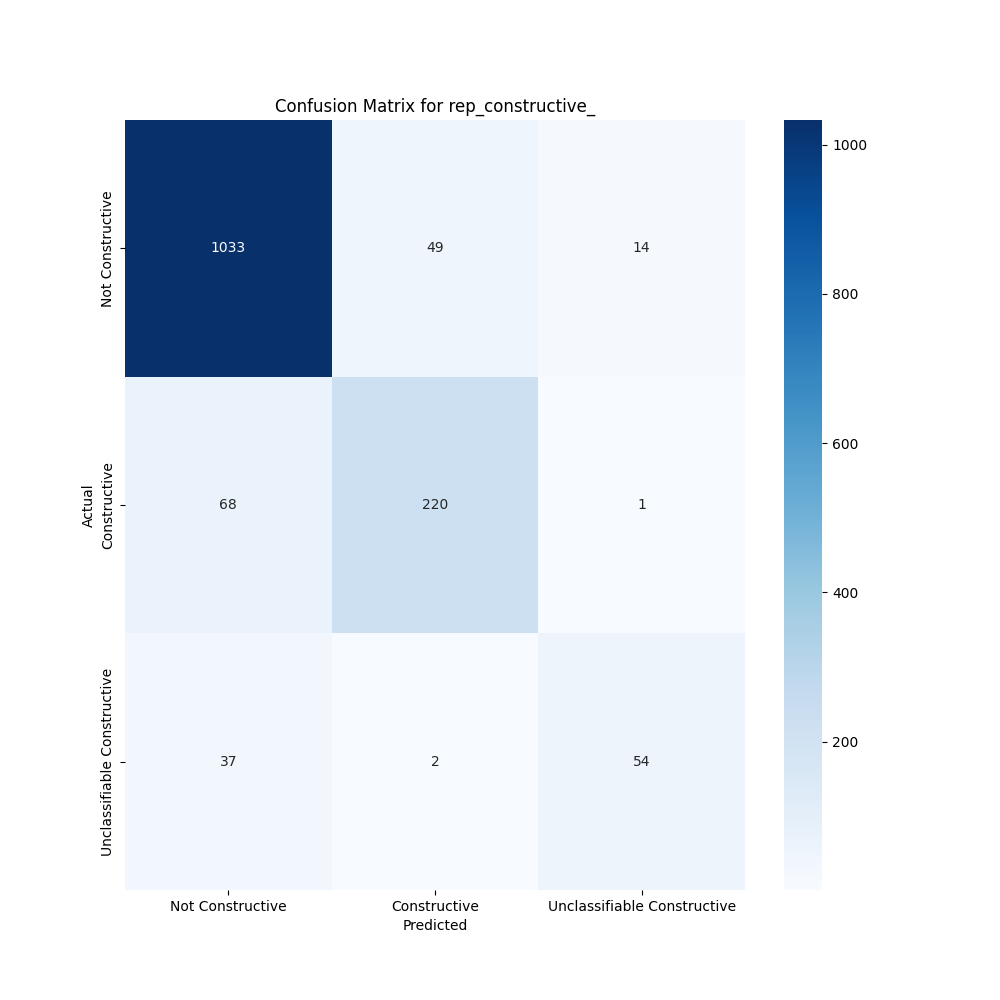}

    \vspace{0.2cm}

    \includegraphics[width=0.85\linewidth]{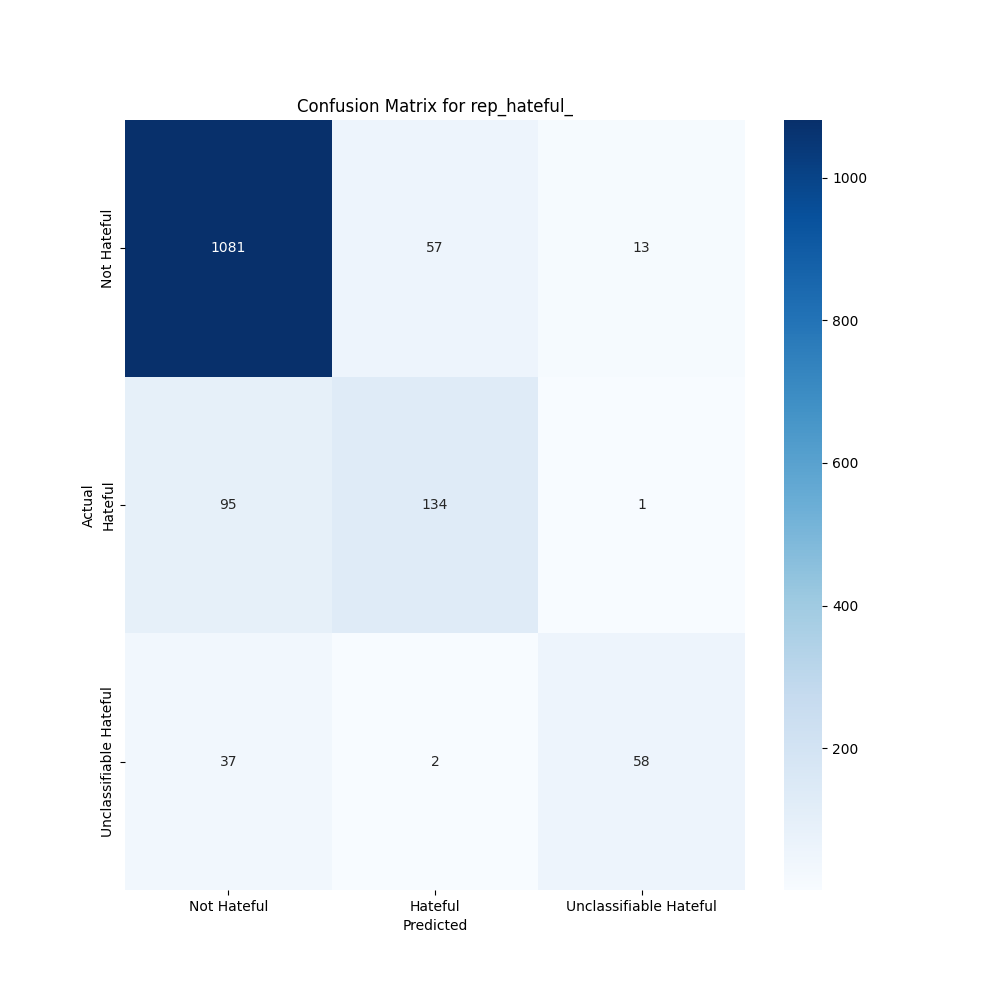}

    \caption{Confusion matrices for the non-context-dependent XLM-RoBERTa-large model: (a) agreement, (b) constructive speech, and (c) hateful speech.}
    \label{fig:xlm-noncon-confusion}
\end{figure}

\subsection{multilingual-e5-large}

\begin{figure}[H]
    \centering

    \includegraphics[width=0.85\linewidth]{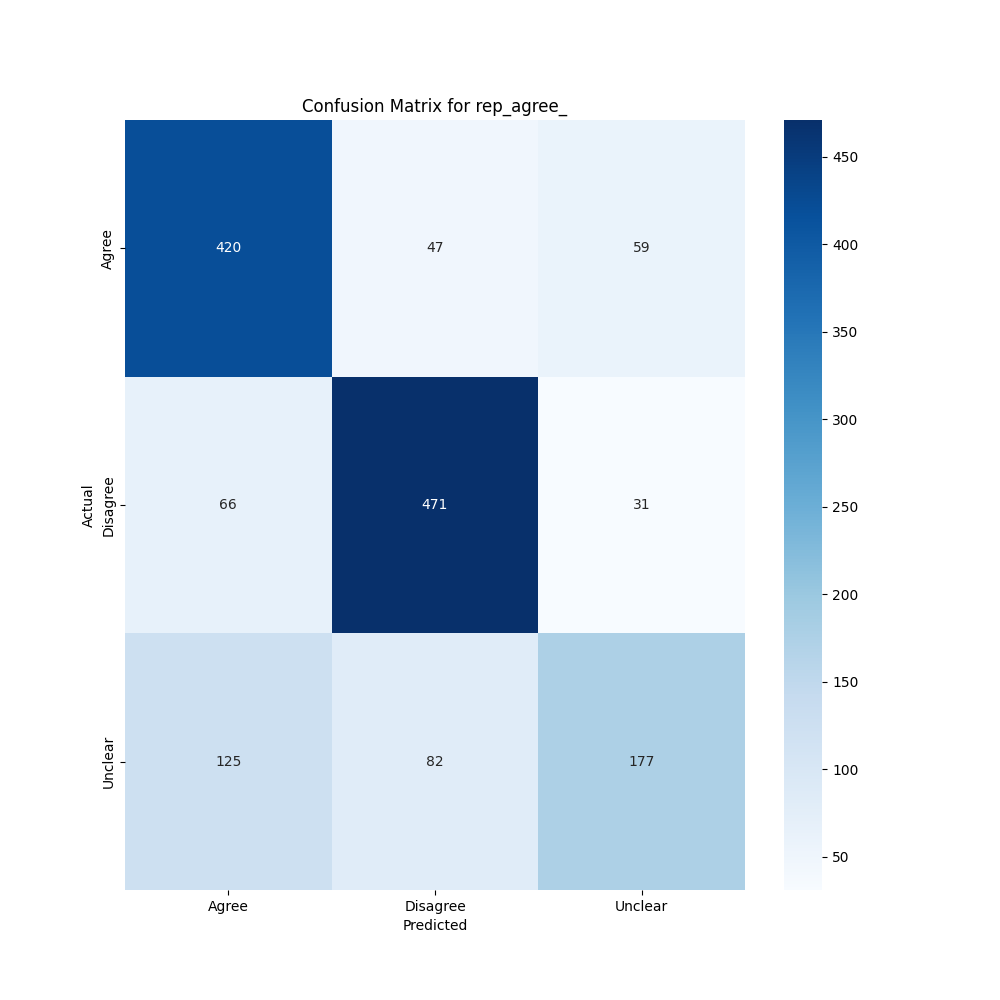}

    \vspace{0.2cm}

    \includegraphics[width=0.85\linewidth]{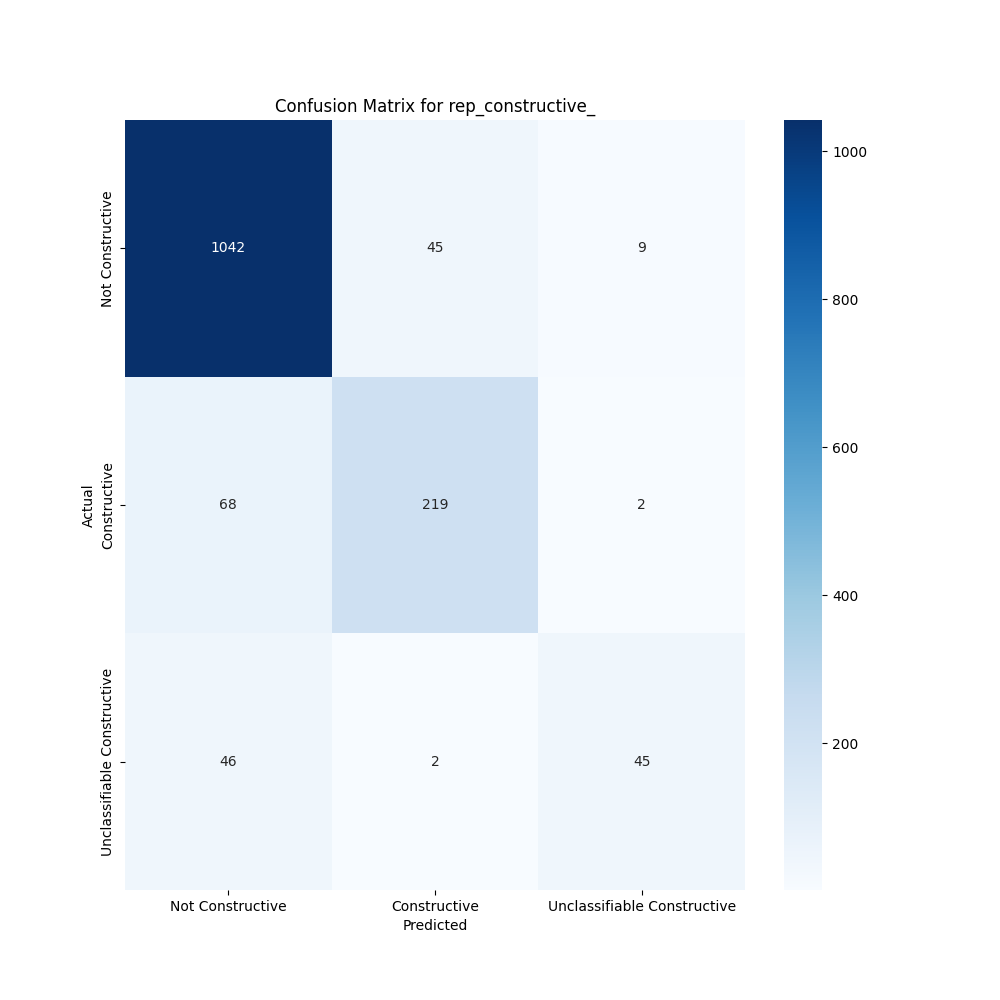}

    \vspace{0.2cm}

    \includegraphics[width=0.85\linewidth]{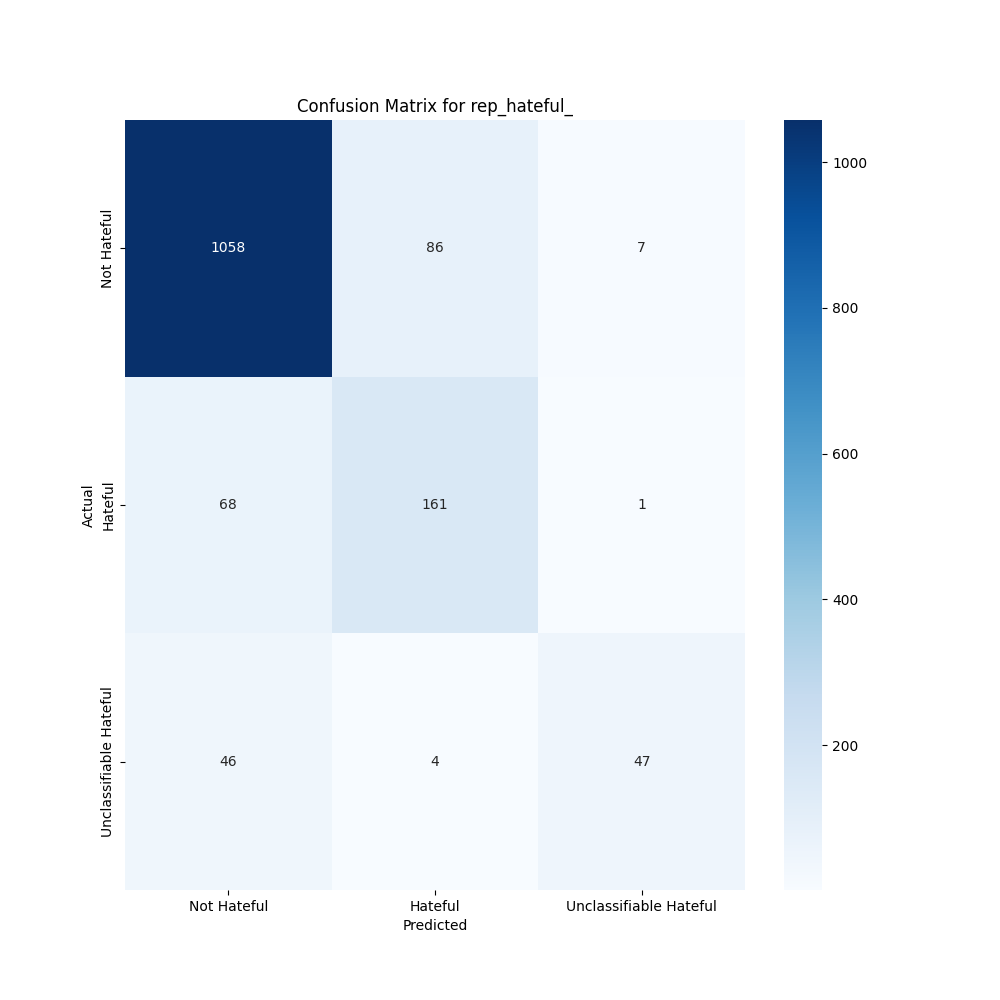}

    \caption{Confusion matrices for the context-dependent multilingual-e5-large model: (a) agreement, (b) constructive speech, and (c) hateful speech.}
    \label{fig:e5-con-confusion}
\end{figure}

\begin{figure}[H]
    \centering

    \includegraphics[width=0.85\linewidth]{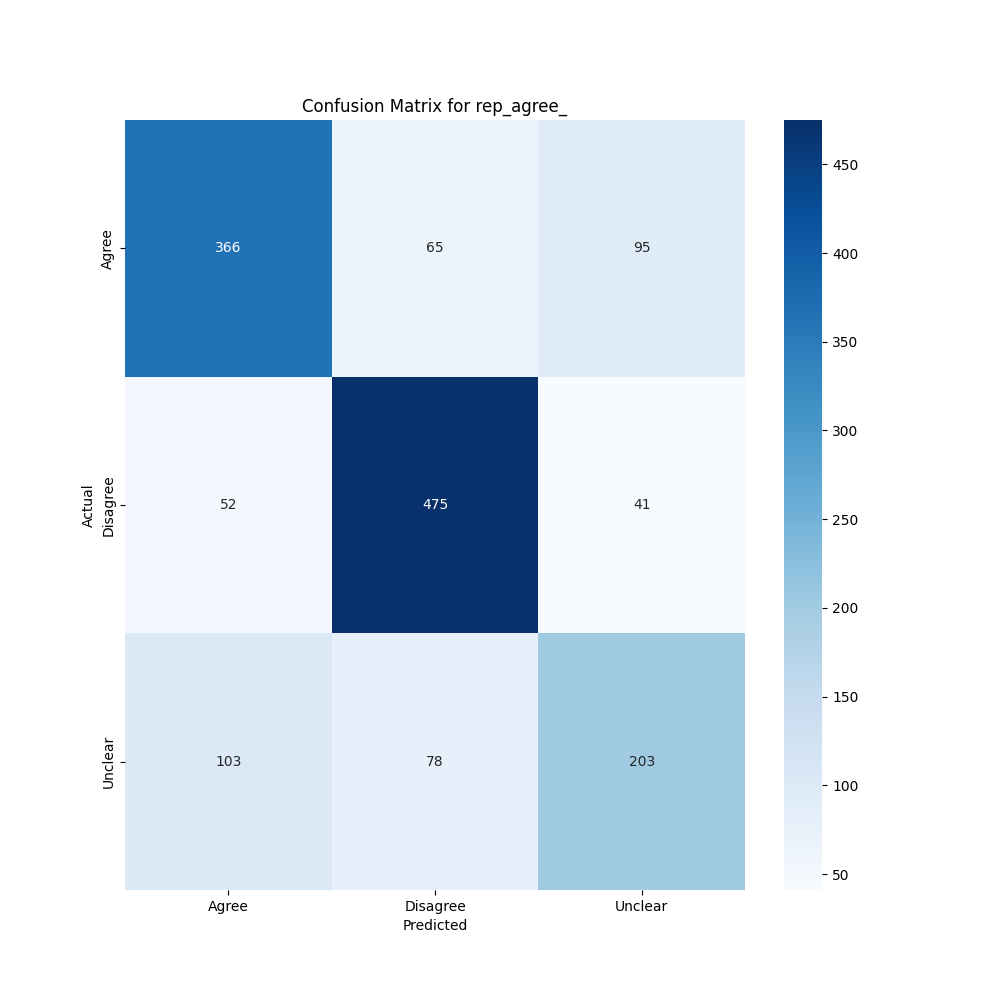}

    \vspace{0.2cm}

    \includegraphics[width=0.85\linewidth]{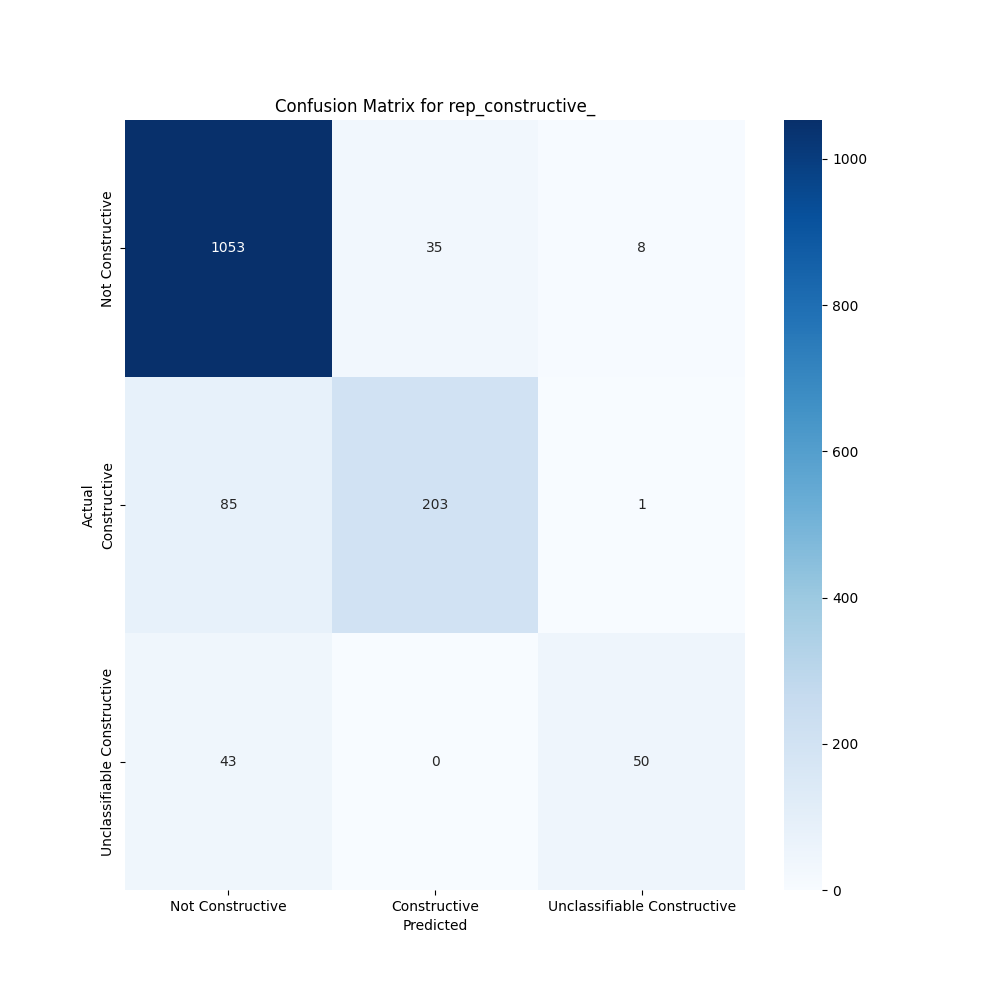}

    \vspace{0.2cm}

    \includegraphics[width=0.85\linewidth]{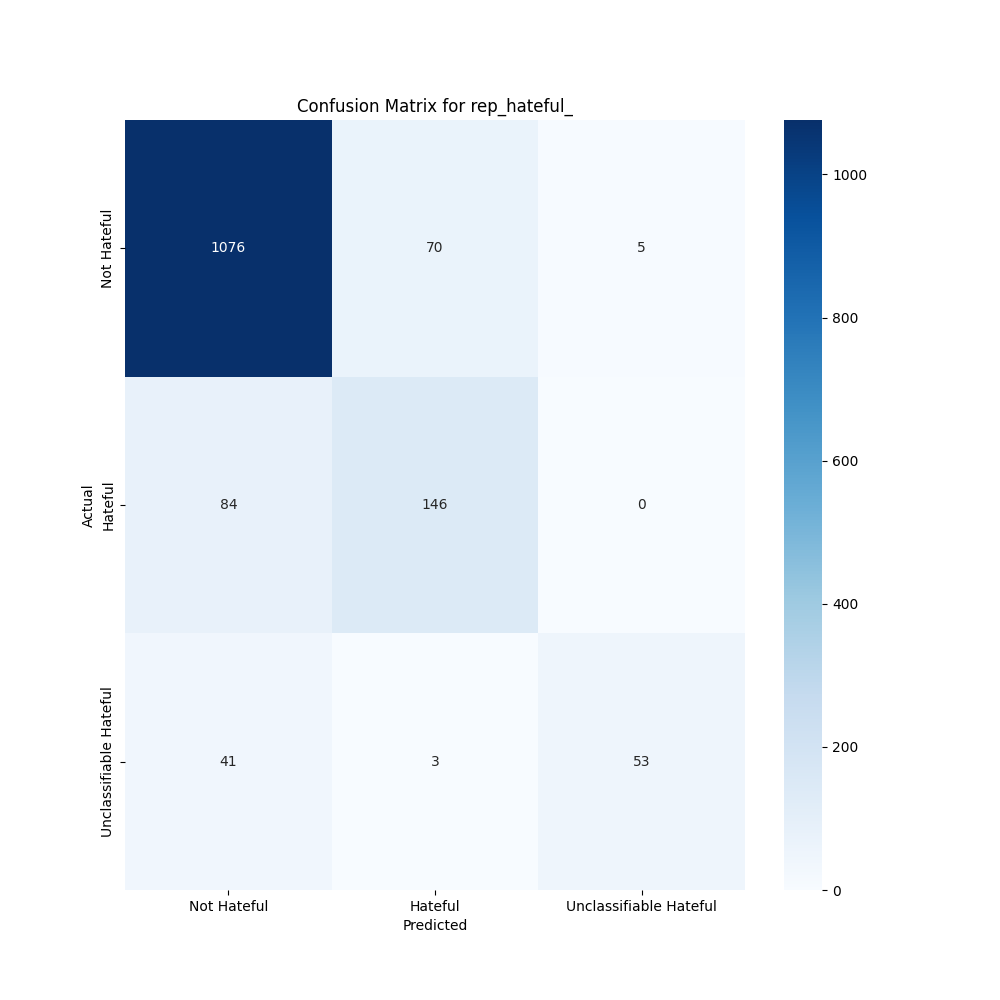}

    \caption{Confusion matrices for the non-context-dependent multilingual-e5-large model: (a) agreement, (b) constructive speech, and (c) hateful speech.}
    \label{fig:e5-noncon-confusion}
\end{figure}

\subsection{Twitter-XLM-RoBERTa-base}

\begin{figure}[H]
    \centering

    \includegraphics[width=0.85\linewidth]{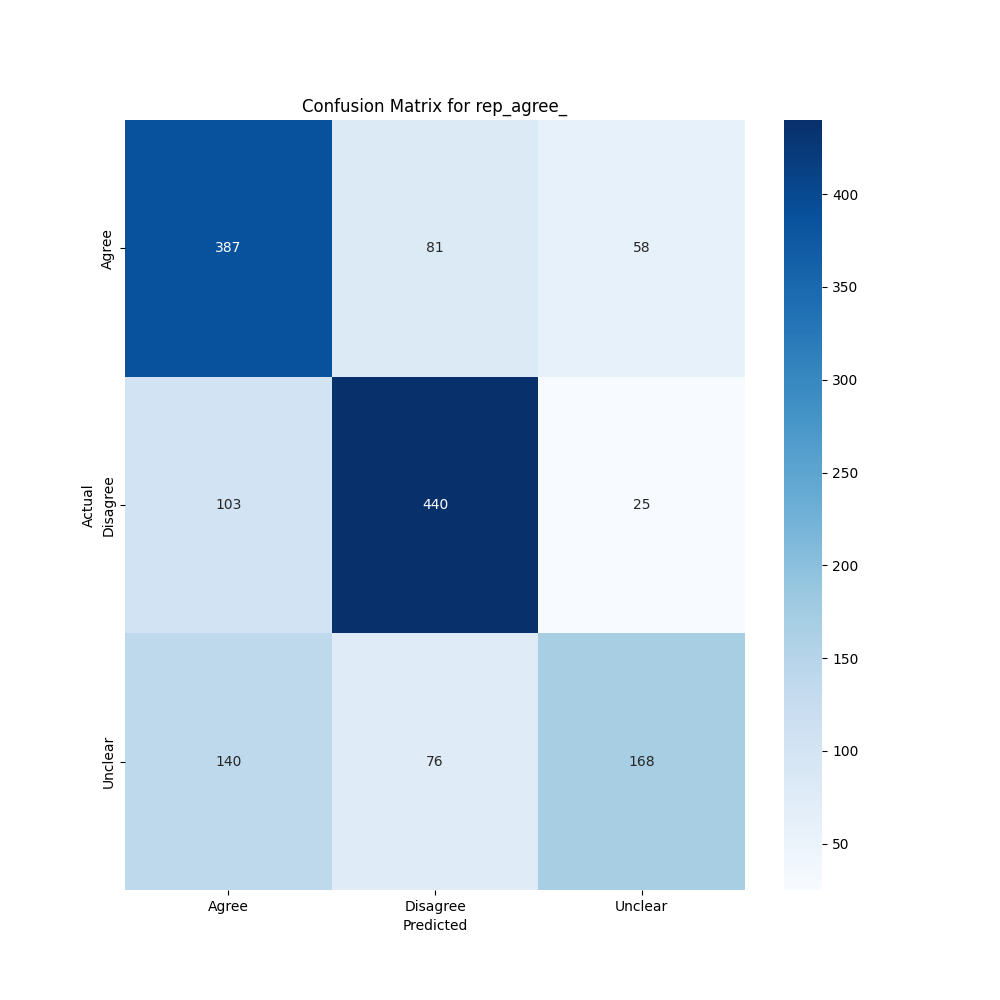}

    \vspace{0.2cm}

    \includegraphics[width=0.85\linewidth]{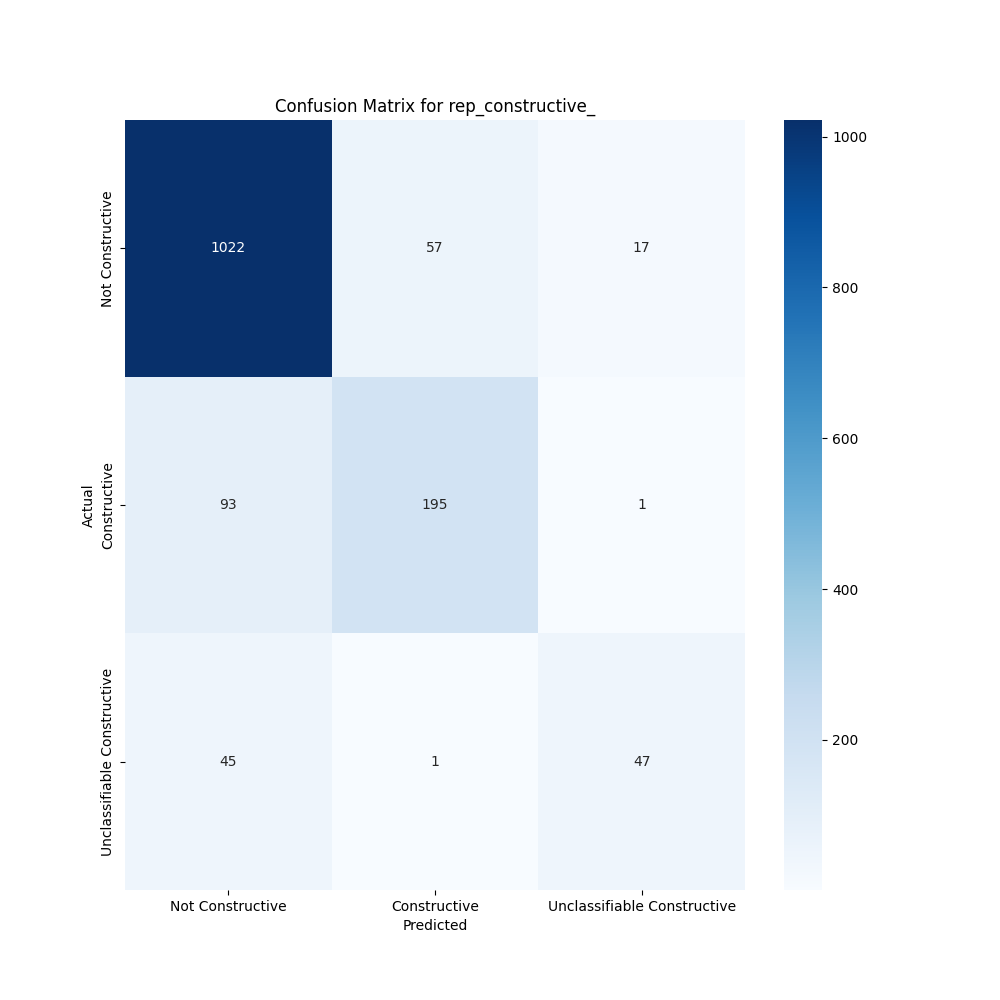}

    \vspace{0.2cm}

    \includegraphics[width=0.85\linewidth]{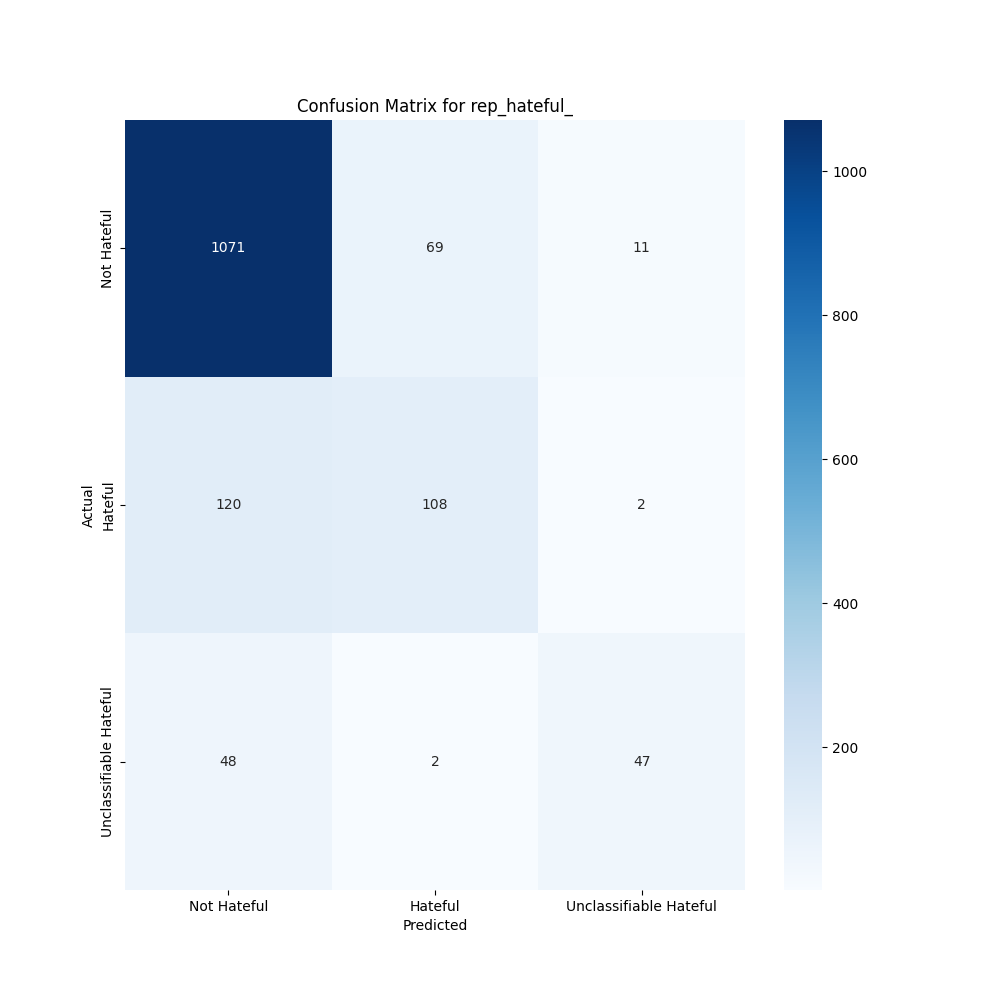}

    \caption{Confusion matrices for the context-dependent Twitter-XLM-RoBERTa-base model: (a) agreement, (b) constructive speech, and (c) hateful speech.}
    \label{fig:twitter-con-confusion}
\end{figure}

\begin{figure}[H]
    \centering

    \includegraphics[width=0.85\linewidth]{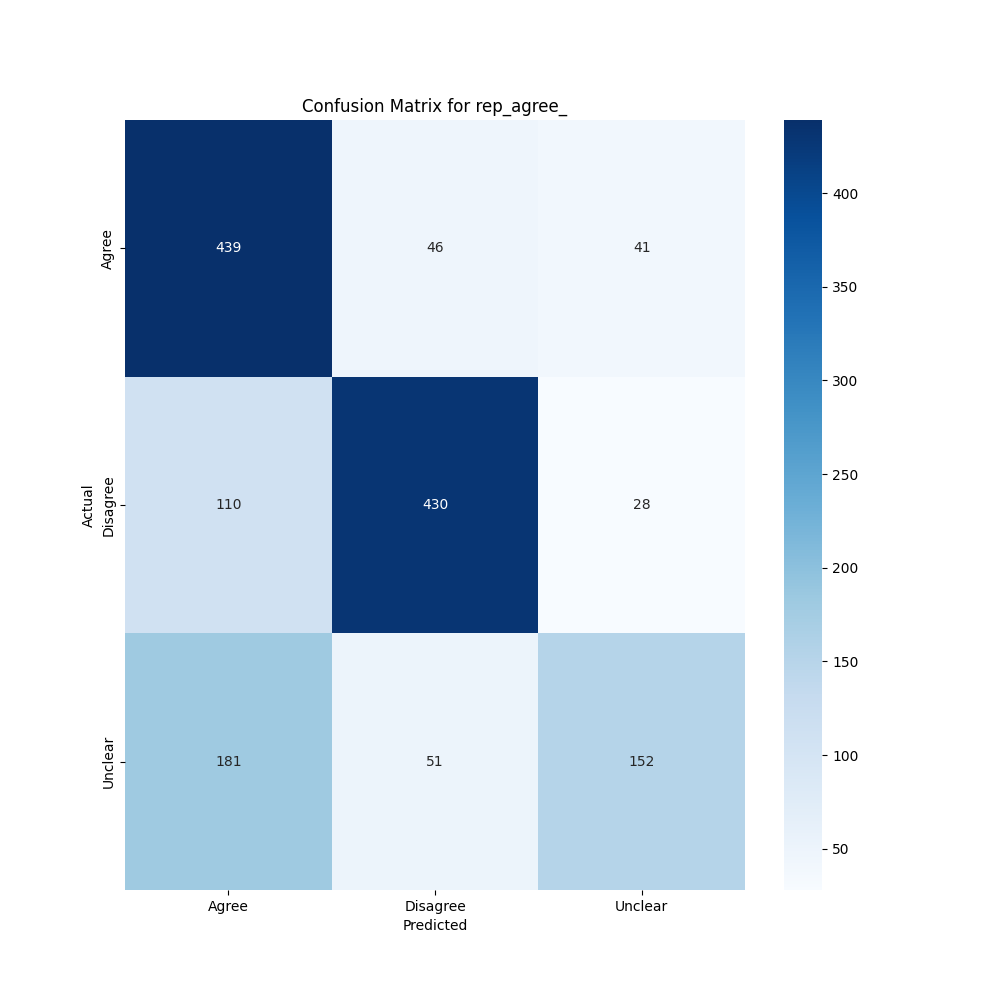}

    \vspace{0.2cm}

    \includegraphics[width=0.85\linewidth]{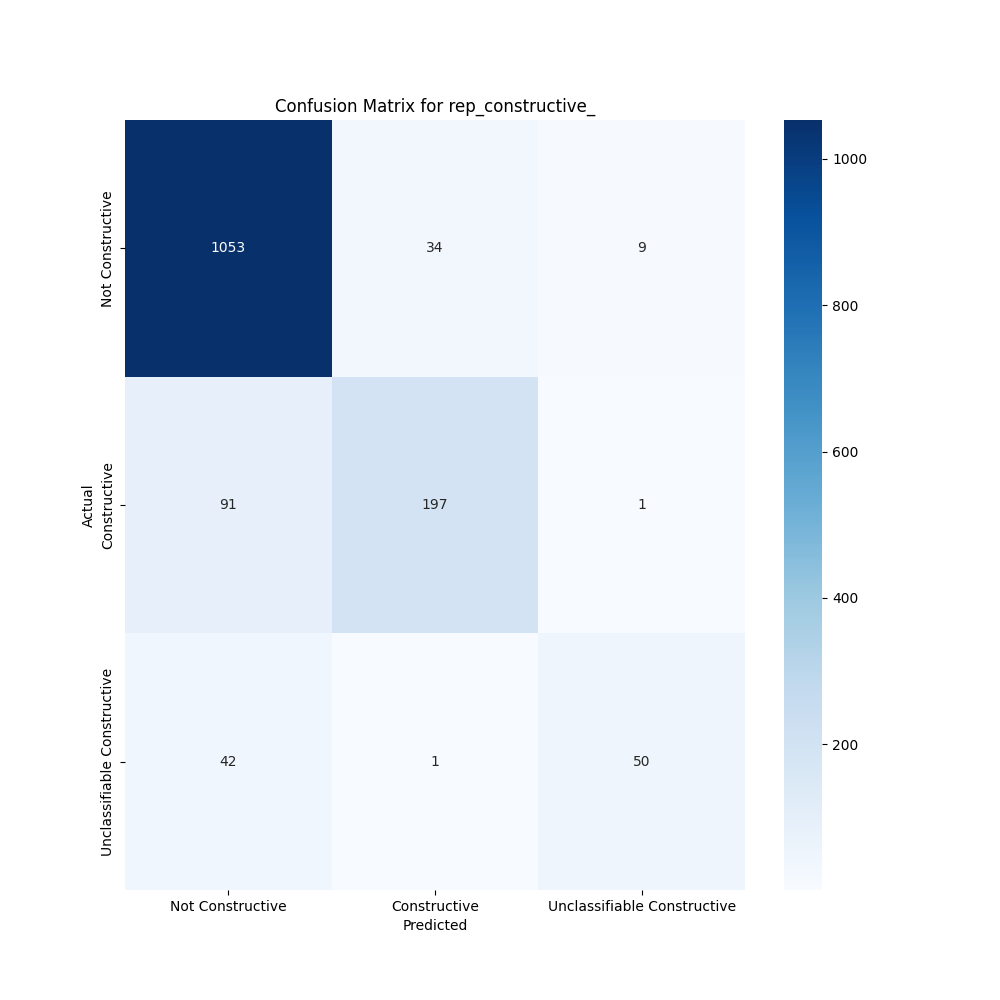}

    \vspace{0.2cm}

    \includegraphics[width=0.85\linewidth]{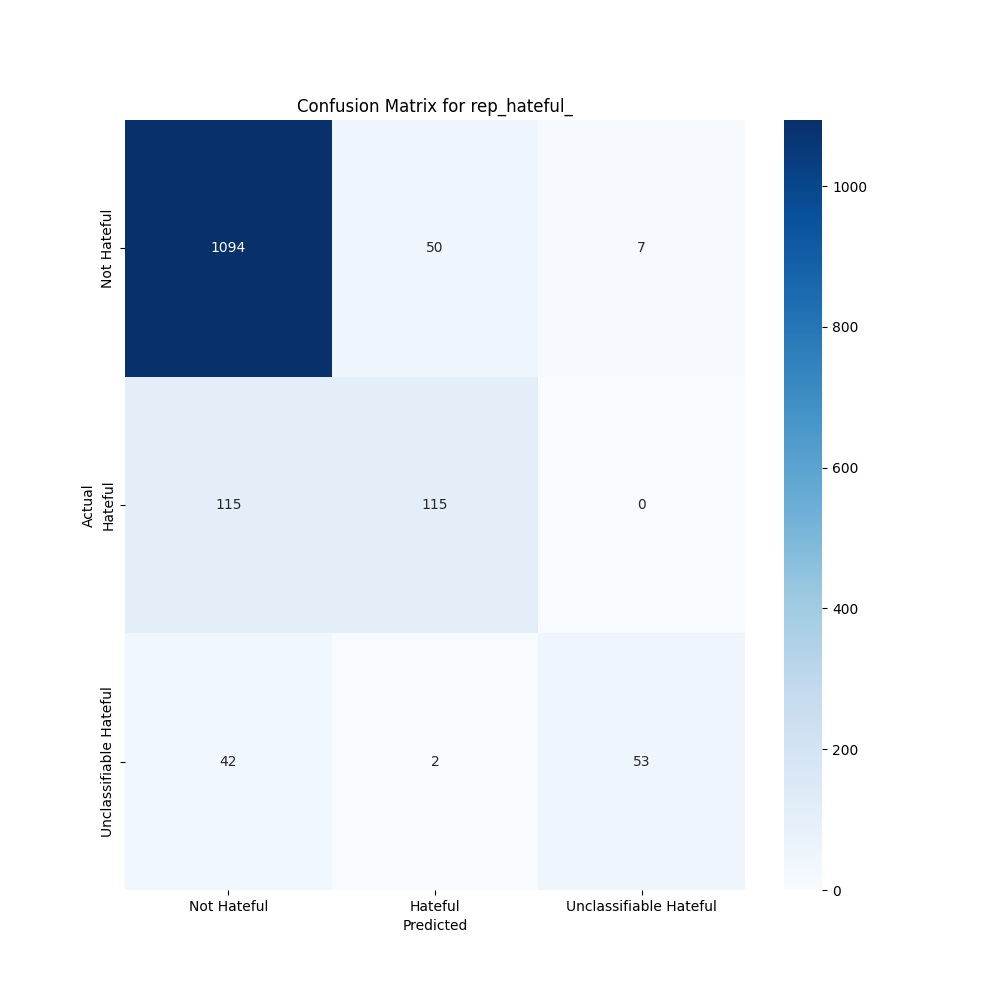}

    \caption{Confusion matrices for the non-context-dependent Twitter-XLM-RoBERTa-base model: (a) agreement, (b) constructive speech, and (c) hateful speech.}
    \label{fig:twitter-noncon-confusion}
\end{figure}

\subsection{mmBERT-base}

\begin{figure}[H]
    \centering

    \includegraphics[width=0.85\linewidth]{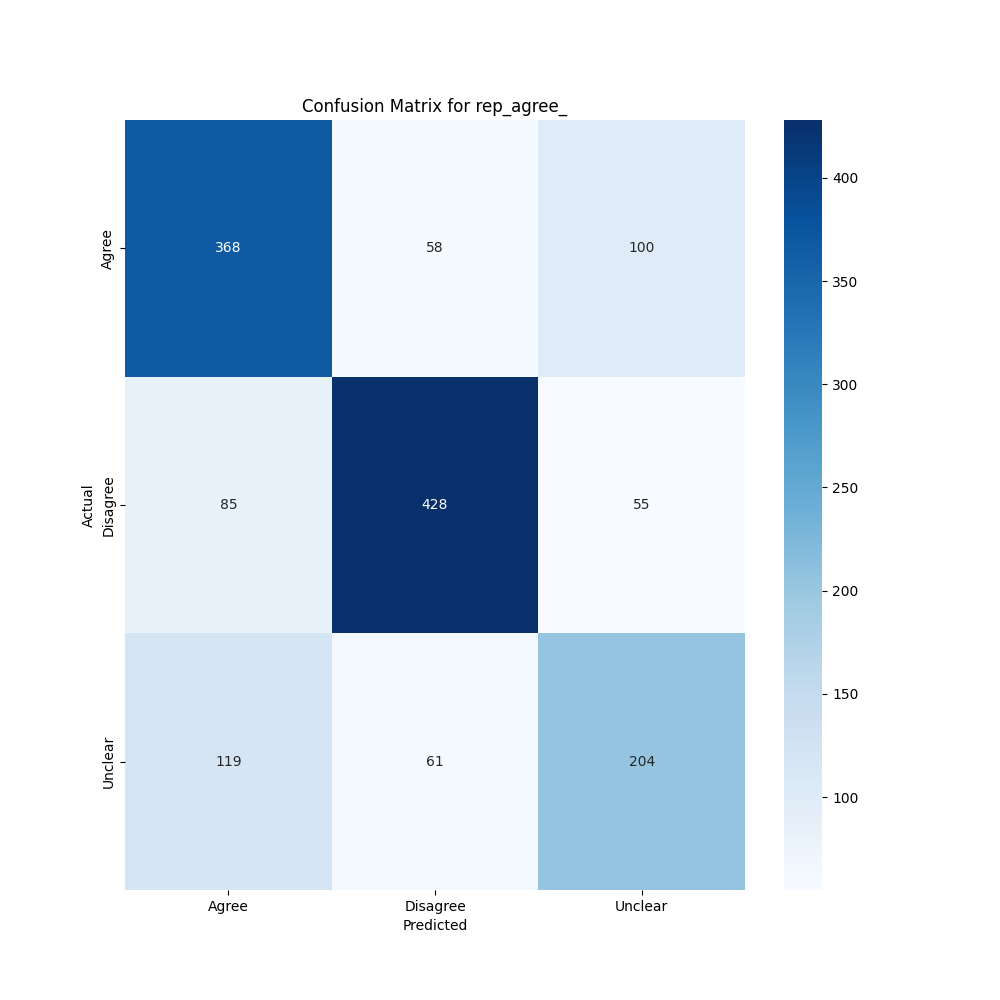}

    \vspace{0.2cm}

    \includegraphics[width=0.85\linewidth]{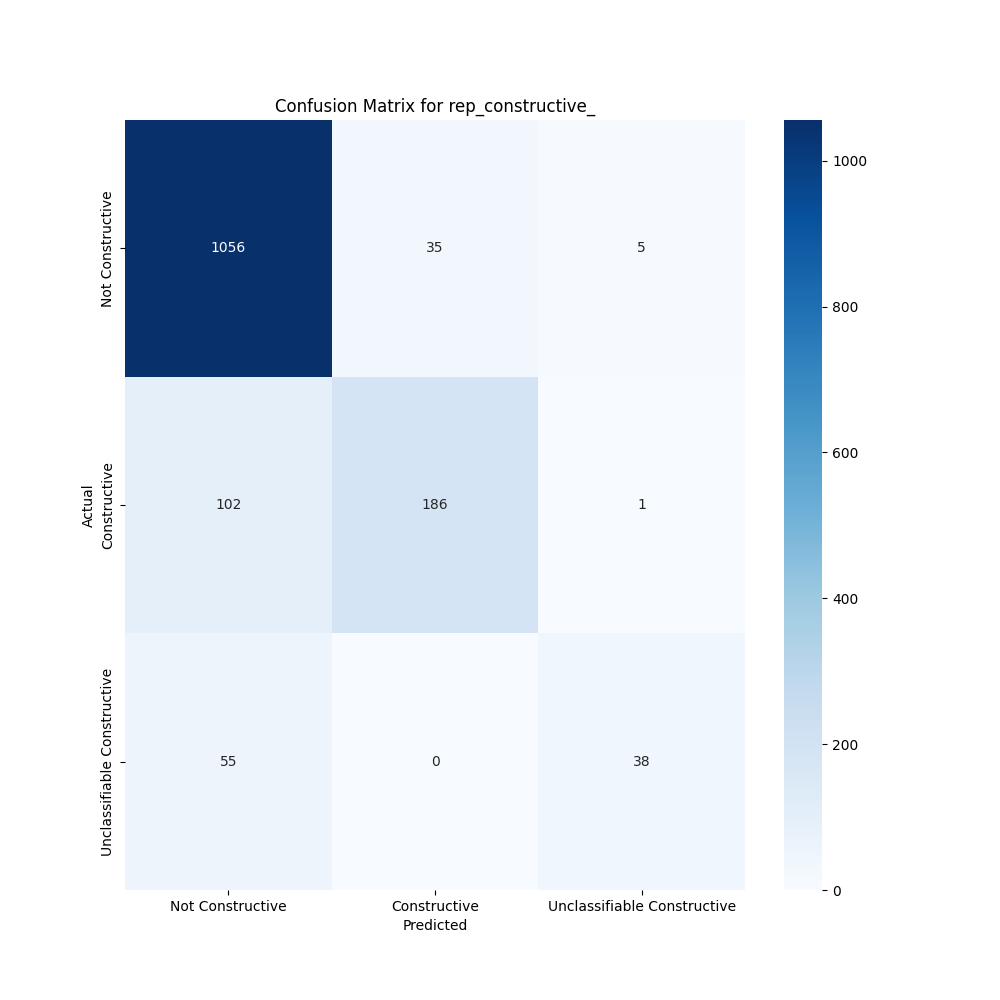}

    \vspace{0.2cm}

    \includegraphics[width=0.85\linewidth]{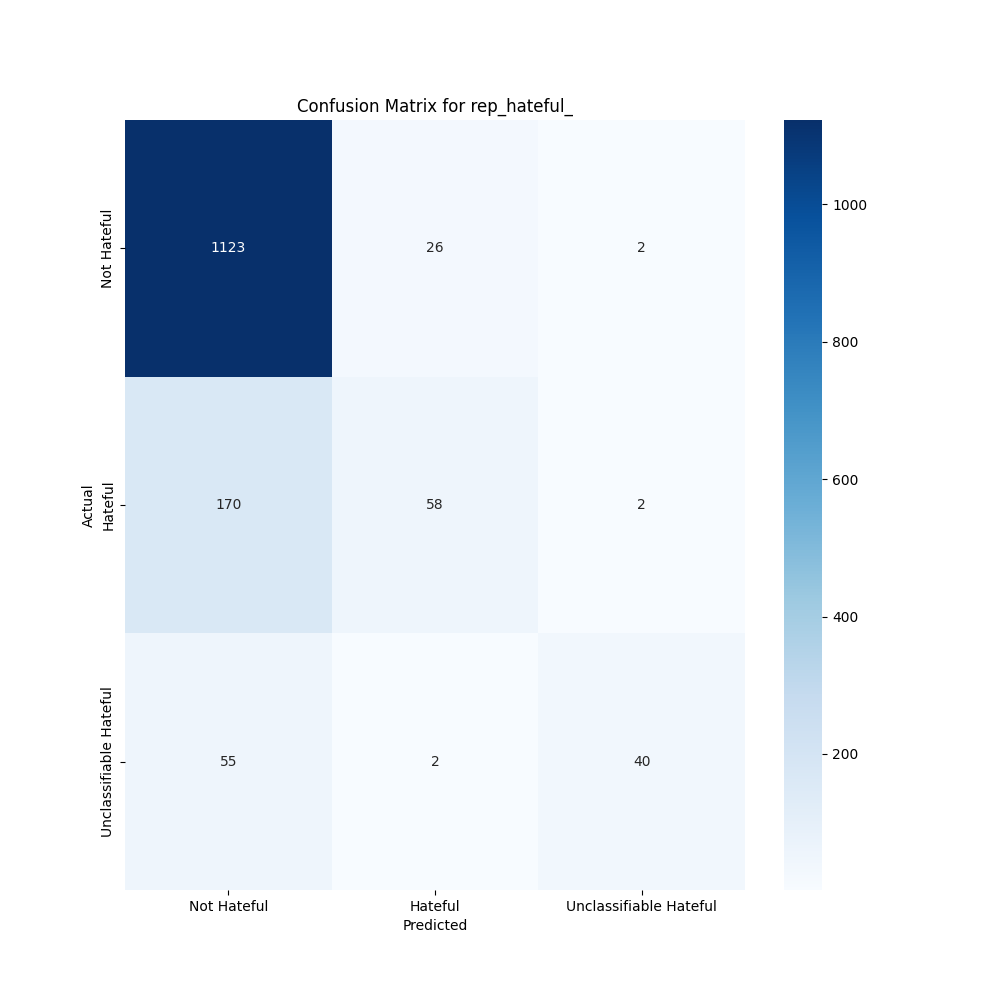}

    \caption{Confusion matrices for the context-dependent mmBERT-base model: (a) agreement, (b) constructive speech, and (c) hateful speech.}
    \label{fig:mmbert-con-confusion}
\end{figure}

\begin{figure}[H]
    \centering

    \includegraphics[width=0.85\linewidth]{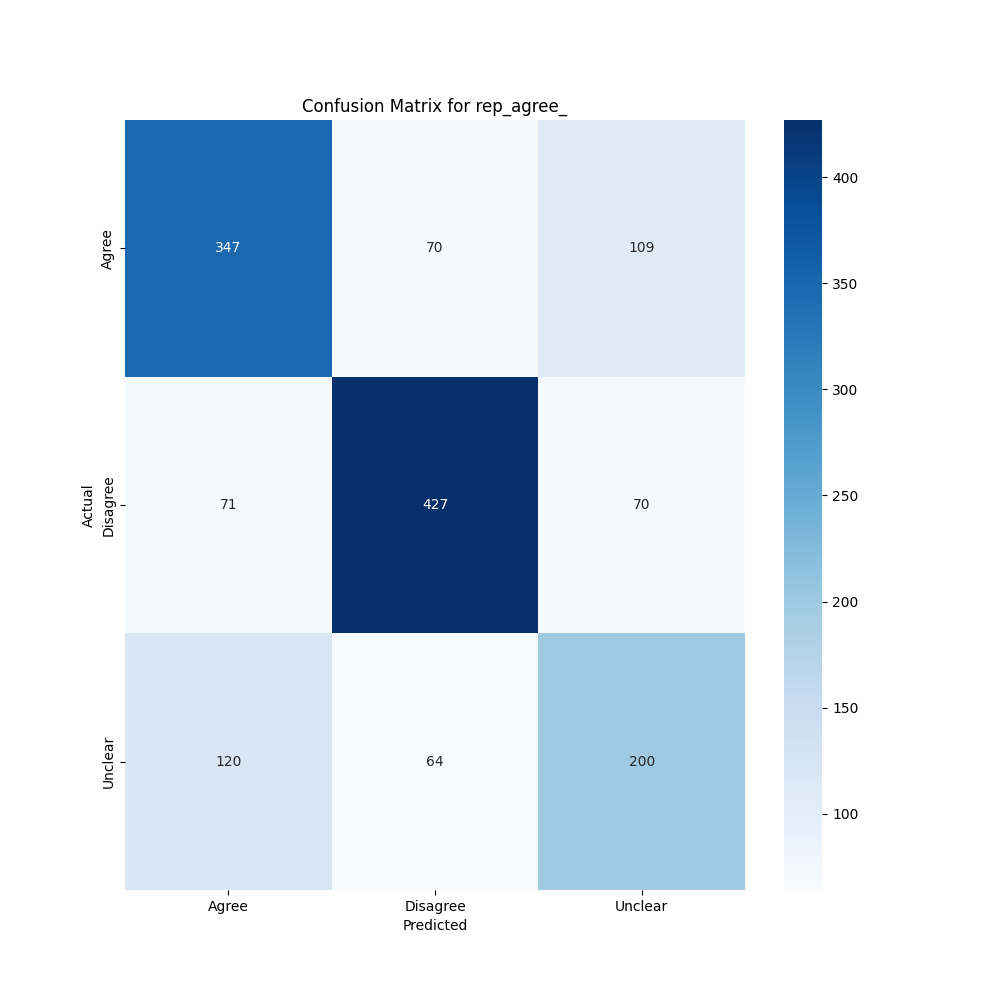}

    \vspace{0.2cm}

    \includegraphics[width=0.85\linewidth]{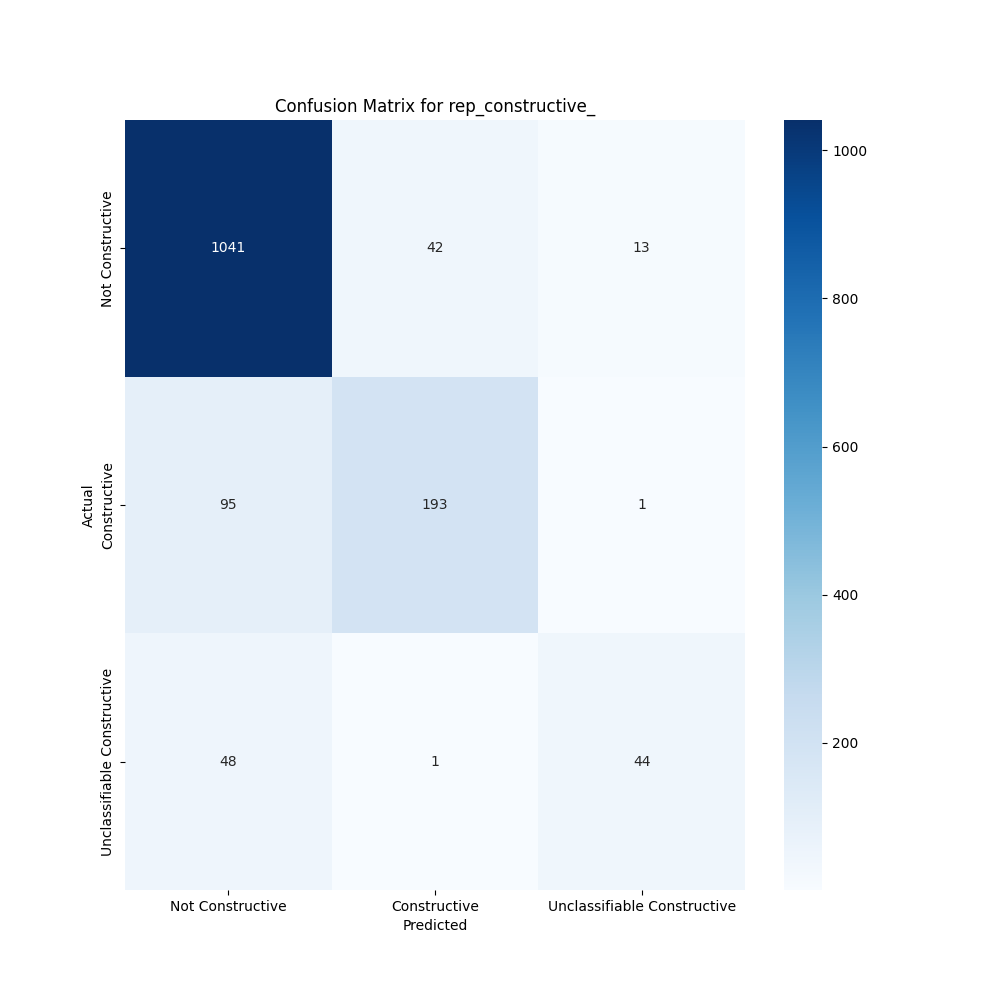}

    \vspace{0.2cm}

    \includegraphics[width=0.85\linewidth]{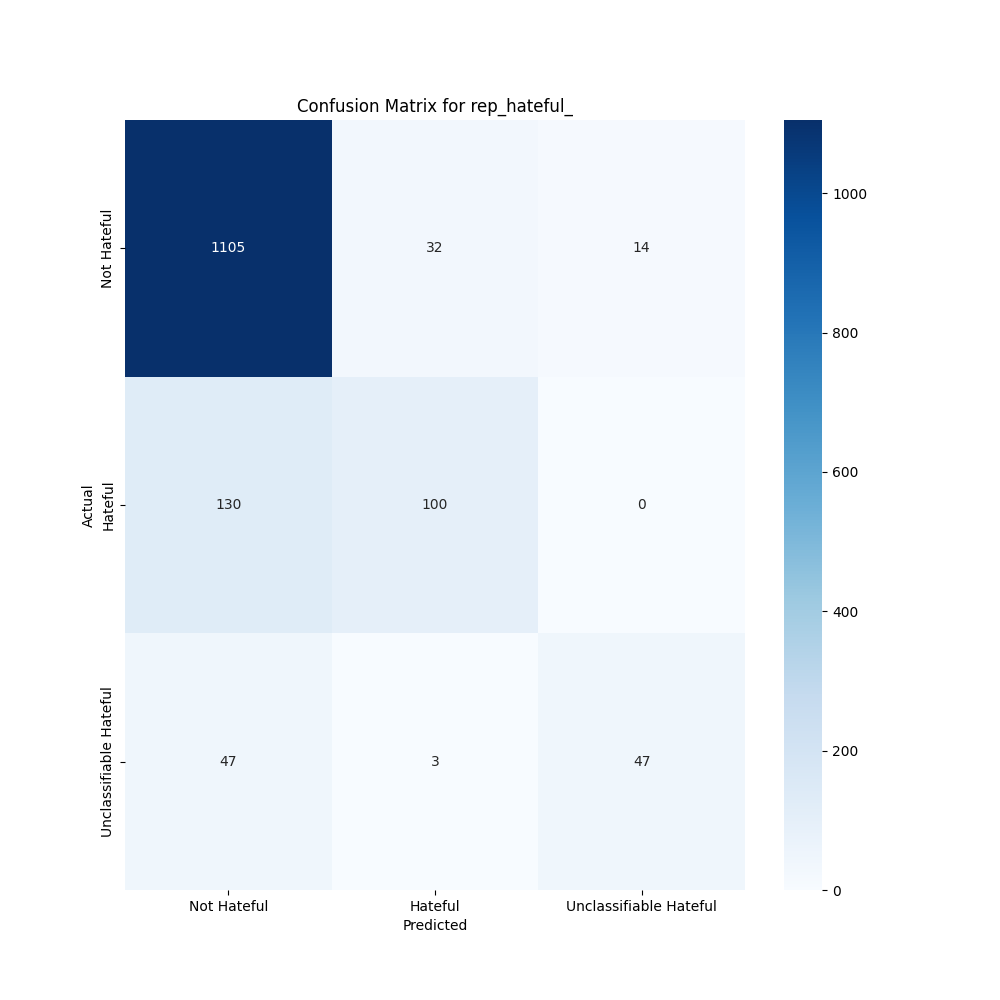}

    \caption{Confusion matrices for the non-context-dependent mmBERT-base model: (a) agreement, (b) constructive speech, and (c) hateful speech.}
    \label{fig:mmbert-noncon-confusion}
\end{figure}

\clearpage
\section{Appendix: Benchmark test against Jev}
\label{sec:appendix}

\begin{table*}[!b]
\centering
\small
\begin{tabular}{@{}l p{8.6cm} p{4.6cm}@{}}
\hline
\textbf{Dimension} & \textbf{Instruction} & \textbf{Classes (criteria)} \\
\hline
Constructive &
Does the \texttt{reply} contain constructive or prosocial elements or does it agree with such elements contained in \texttt{reference}? Constructive elements signal the disapproval of hate or hostile behavior, or side with, shield, or comfort an individual victim or victimized group. &
\textit{constructive}: The \texttt{reply} is constructive or prosocial; 
\textit{not constructive}: The \texttt{reply} is NOT constructive or prosocial;
\textit{unclassifiable}: The constructiveness of \texttt{reply} is not classifiable because it is too ambiguous, or its meaning otherwise unclear\\
\hline
Hateful &
Does the \texttt{reply} contain hateful elements or does it agree with hateful elements contained in \texttt{reference}? Hateful elements are defined as any expression of the motivation to degrade, harass, or intimidate individuals or collectives based on political differences or discriminatory views. &
\textit{hateful}: The \texttt{reply} is hateful; 
\textit{not hateful}: The \texttt{reply} is NOT hateful; 
\textit{unclassifiable}: The hatefulness of \texttt{reply} is not classifiable because it is too ambiguous, or its meaning otherwise unclear\\
\hline
Agreement &
Does the author of \texttt{reply} agree in spirit with what is said in \texttt{reference}? &
\textit{agree}: The author agrees;
\textit{disagree}: The author disagrees; 
\textit{unclear}: It is unclear whether the author agrees or disagrees\\
\hline
\end{tabular}
\caption{Jev specification. Each question received the \texttt{reference}--\texttt{reply} pair as input state; the instruction and the class definitions (criteria) were the only task-specific input.}
\label{tab:jev-spec}
\end{table*}

To evaluate our model performance against recent, state of the art classifier models, we used Jev (version 1.13) \citep{typesafe2026jev} to classify all of our model training data on the dimensions outlined above (hateful, constructive, agreement). We settled on Jev since it is fast, cheap, works as multilabel classifier, and can be used as a zero-shot classifier, thus promising to provide a classification infrastructure at scale without the need for expensive training data.

We ran Jev on 14 October 2026 on all 7{,}388 reply--reference pairs (6{,}199 English, 1{,}189 Danish). Each pair was passed to Jev as a JSON object with two fields, \texttt{reference} (the original post) and \texttt{reply}, mimicking the context-enriched scenario, together with three independent choice questions, one per dimension (Table~\ref{tab:jev-spec}).
\newpage
Each choice question consists of an \emph{instruction}, which poses the annotation question and refers to the two fields by name, and \emph{criteria}, which define the possible answers in plain language. As in the human annotation, every question has a third, residual class (\textit{unclassifiable} or \textit{unclear}). The questions were sent in a single request, answered independently of one another, and we took the class with the highest probability as Jev's prediction. The English instructions were used unchanged for both languages. We provided no labelled examples and did no prompt tuning, and instead relied on the codebook for task definitions.

Since performance differed substantially for English and Danish data, we provide performance metrics for combined and English / Danish separately.

\clearpage
\raggedbottom

\subsection{Overall performance (English + Danish)}

\begin{table}[H]
\centering
\small
\setlength{\tabcolsep}{4pt}
\begin{tabular}{lrrrr}
\hline
\textbf{Class} & \textbf{P} & \textbf{R} & \textbf{F1} & \textbf{Support} \\
\hline
\multicolumn{5}{l}{\textbf{Constructive}} \\
Constructive & 0.505 & 0.881 & 0.642 & 1,441 \\
Not constructive & 0.917 & 0.768 & 0.836 & 5,465 \\
Unclassifiable & 0.675 & 0.413 & 0.512 & 482 \\
\textit{Macro avg} & \textit{0.699} & \textit{0.688} & \textit{0.663} & \textit{7,388} \\
\textit{Weighted avg} & \textit{0.821} & \textit{0.767} & \textit{0.777} & \textit{7,388} \\
\hline
\multicolumn{5}{l}{\textbf{Hateful}} \\
Hateful & 0.560 & 0.671 & 0.611 & 1,134 \\
Not hateful & 0.888 & 0.892 & 0.890 & 5,779 \\
Unclassifiable & 0.761 & 0.356 & 0.485 & 475 \\
\textit{Macro avg} & \textit{0.736} & \textit{0.640} & \textit{0.662} & \textit{7,388} \\
\textit{Weighted avg} & \textit{0.829} & \textit{0.824} & \textit{0.821} & \textit{7,388} \\
\hline
\multicolumn{5}{l}{\textbf{Agreement}} \\
Agree & 0.700 & 0.815 & 0.753 & 2,711 \\
Disagree & 0.731 & 0.840 & 0.782 & 2,811 \\
Unclear & 0.673 & 0.361 & 0.470 & 1,866 \\
\textit{Macro avg} & \textit{0.701} & \textit{0.672} & \textit{0.668} & \textit{7,388} \\
\textit{Weighted avg} & \textit{0.705} & \textit{0.710} & \textit{0.693} & \textit{7,388} \\
\hline
\end{tabular}
\caption{Jev vs.\ human annotation, English + Danish ($n = 7{,}388$): precision (P), recall (R) and F1 per class; macro avg.\ is the unweighted mean over the three classes, weighted avg.\ is weighted by support.}
\label{tab:jev-combined}
\end{table}

\begin{figure}[H]
    \centering

    \includegraphics[width=0.85\linewidth]{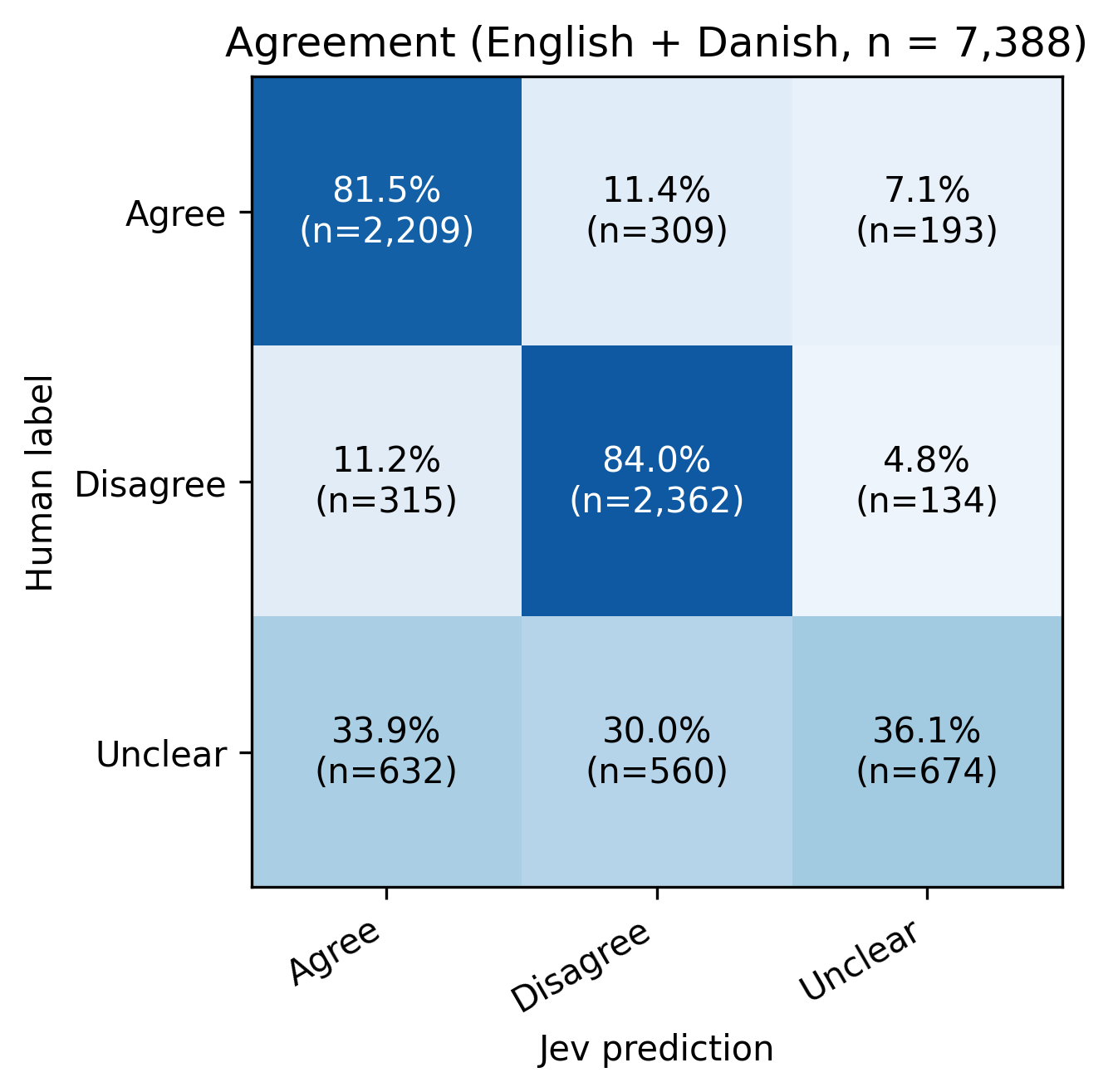}

    \vspace{0.2cm}

    \includegraphics[width=0.85\linewidth]{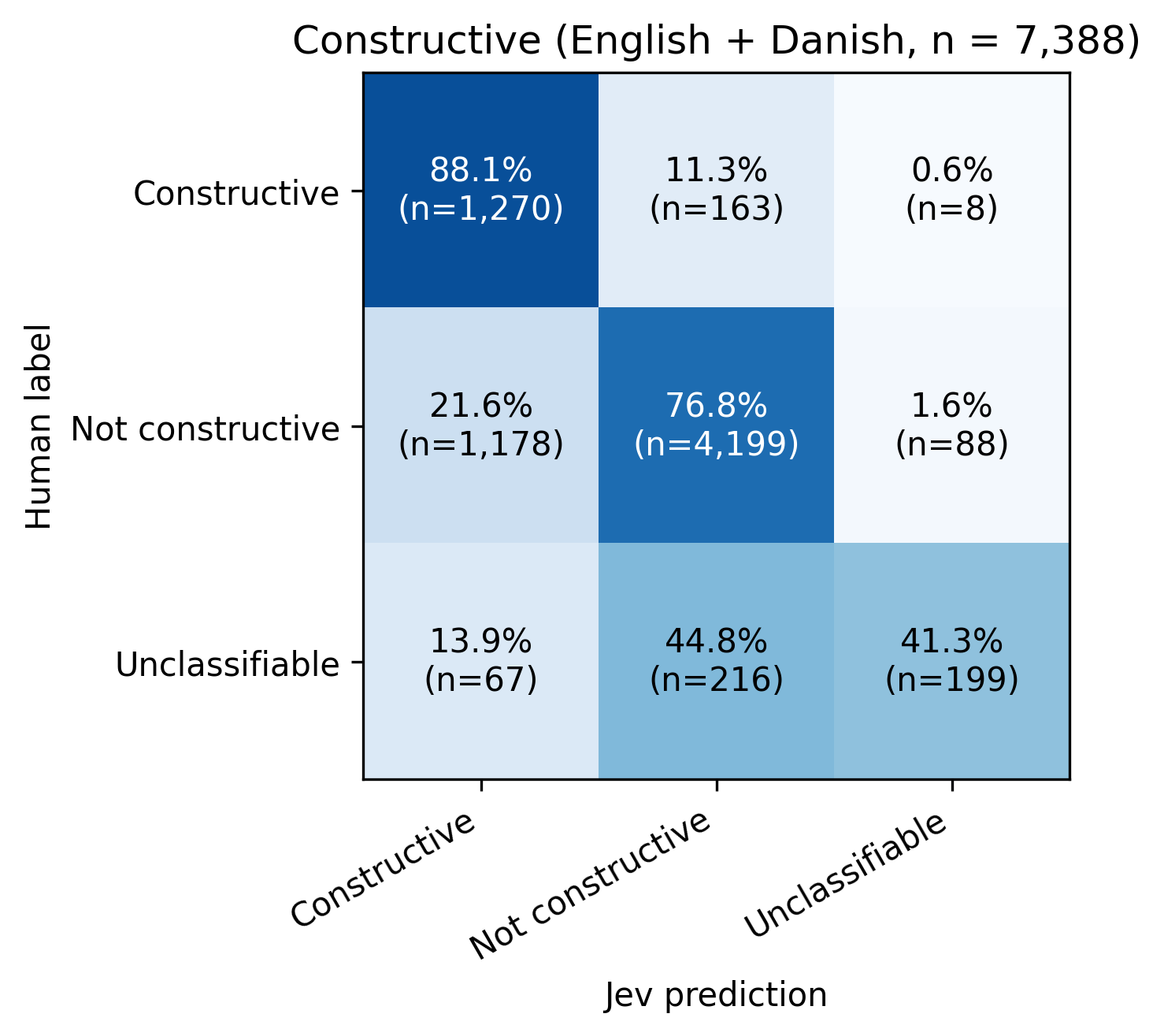}

    \vspace{0.2cm}

    \includegraphics[width=0.85\linewidth]{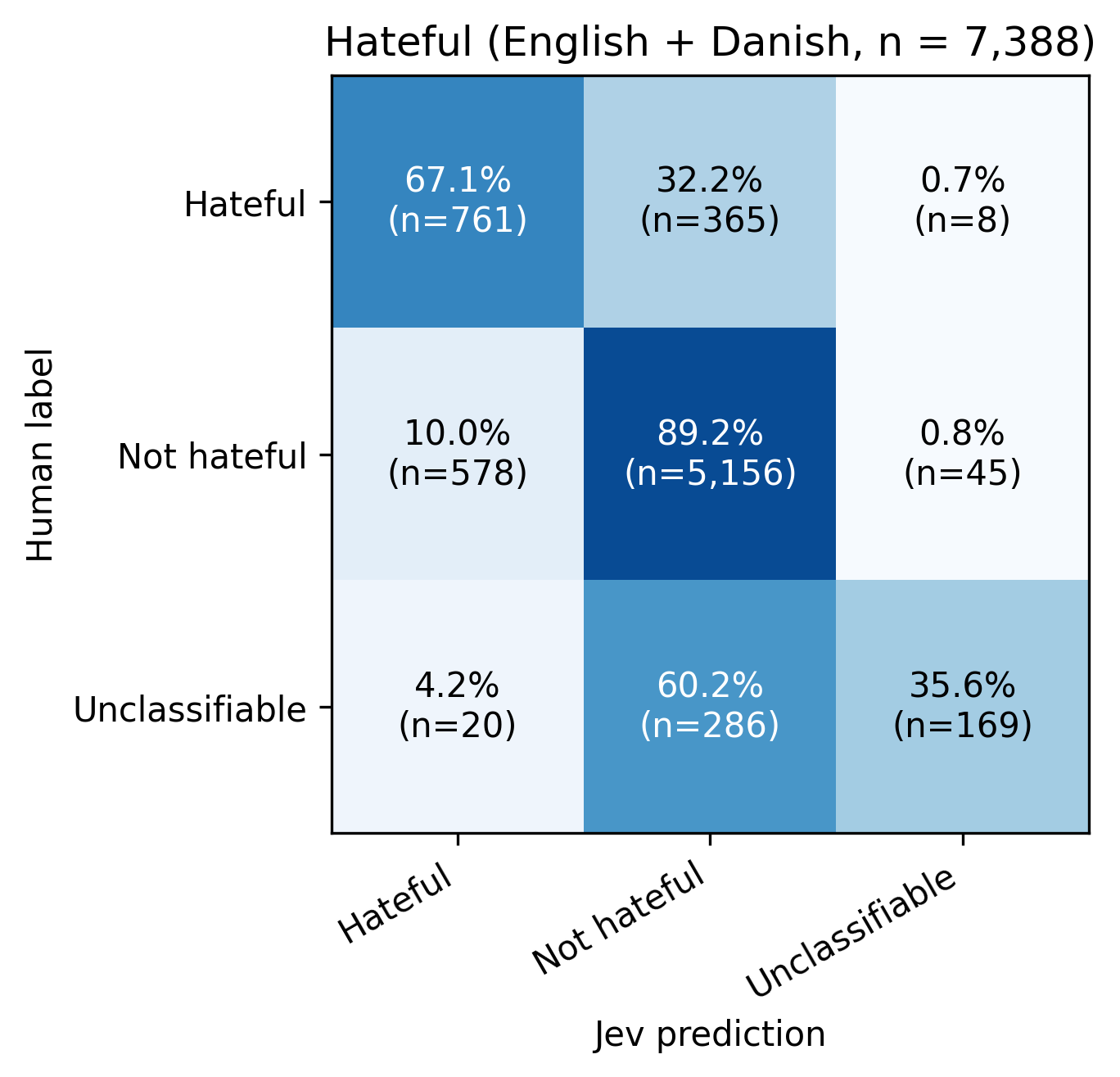}

    \caption{Confusion matrices for the Jev benchmark (English + Danish): (a) agreement, (b) constructive speech, and (c) hateful speech.}
    \label{fig:xlm-confusion}
\end{figure}

\clearpage

\subsection{English performance}

\begin{table}[H]
\centering
\small
\setlength{\tabcolsep}{4pt}
\begin{tabular}{lrrrr}
\hline
\textbf{Class} & \textbf{P} & \textbf{R} & \textbf{F1} & \textbf{Support} \\
\hline
\multicolumn{5}{l}{\textbf{Constructive}} \\
Constructive & 0.570 & 0.906 & 0.700 & 1,281 \\
Not constructive & 0.925 & 0.797 & 0.856 & 4,493 \\
Unclassifiable & 0.679 & 0.468 & 0.554 & 425 \\
\textit{Macro avg} & \textit{0.725} & \textit{0.724} & \textit{0.703} & \textit{6,199} \\
\textit{Weighted avg} & \textit{0.835} & \textit{0.797} & \textit{0.803} & \textit{6,199} \\
\hline
\multicolumn{5}{l}{\textbf{Hateful}} \\
Hateful & 0.549 & 0.694 & 0.613 & 1,009 \\
Not hateful & 0.887 & 0.874 & 0.881 & 4,772 \\
Unclassifiable & 0.761 & 0.404 & 0.528 & 418 \\
\textit{Macro avg} & \textit{0.732} & \textit{0.657} & \textit{0.674} & \textit{6,199} \\
\textit{Weighted avg} & \textit{0.824} & \textit{0.813} & \textit{0.813} & \textit{6,199} \\
\hline
\multicolumn{5}{l}{\textbf{Agreement}} \\
Agree & 0.741 & 0.809 & 0.773 & 2,404 \\
Disagree & 0.741 & 0.852 & 0.792 & 2,351 \\
Unclear & 0.653 & 0.395 & 0.493 & 1,444 \\
\textit{Macro avg} & \textit{0.712} & \textit{0.685} & \textit{0.686} & \textit{6,199} \\
\textit{Weighted avg} & \textit{0.721} & \textit{0.729} & \textit{0.715} & \textit{6,199} \\
\hline
\end{tabular}
\caption{Jev vs.\ human annotation, English ($n = 6{,}199$): precision (P), recall (R) and F1 per class; macro avg.\ is the unweighted mean over the three classes, weighted avg.\ is weighted by support.}
\label{tab:jev-en}
\end{table}

\begin{figure}[H]
    \centering

    \includegraphics[width=0.85\linewidth]{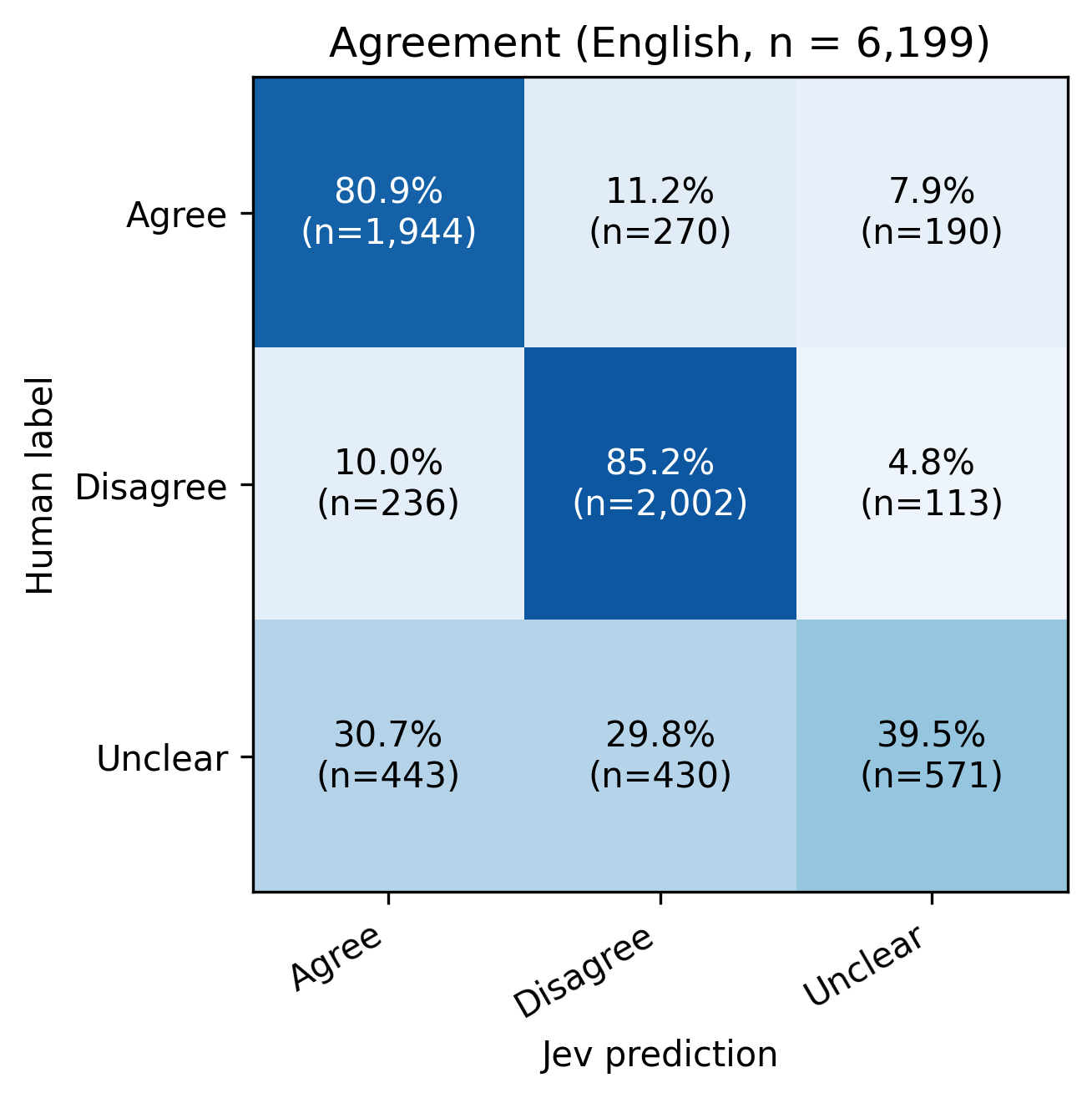}

    \vspace{0.2cm}

    \includegraphics[width=0.85\linewidth]{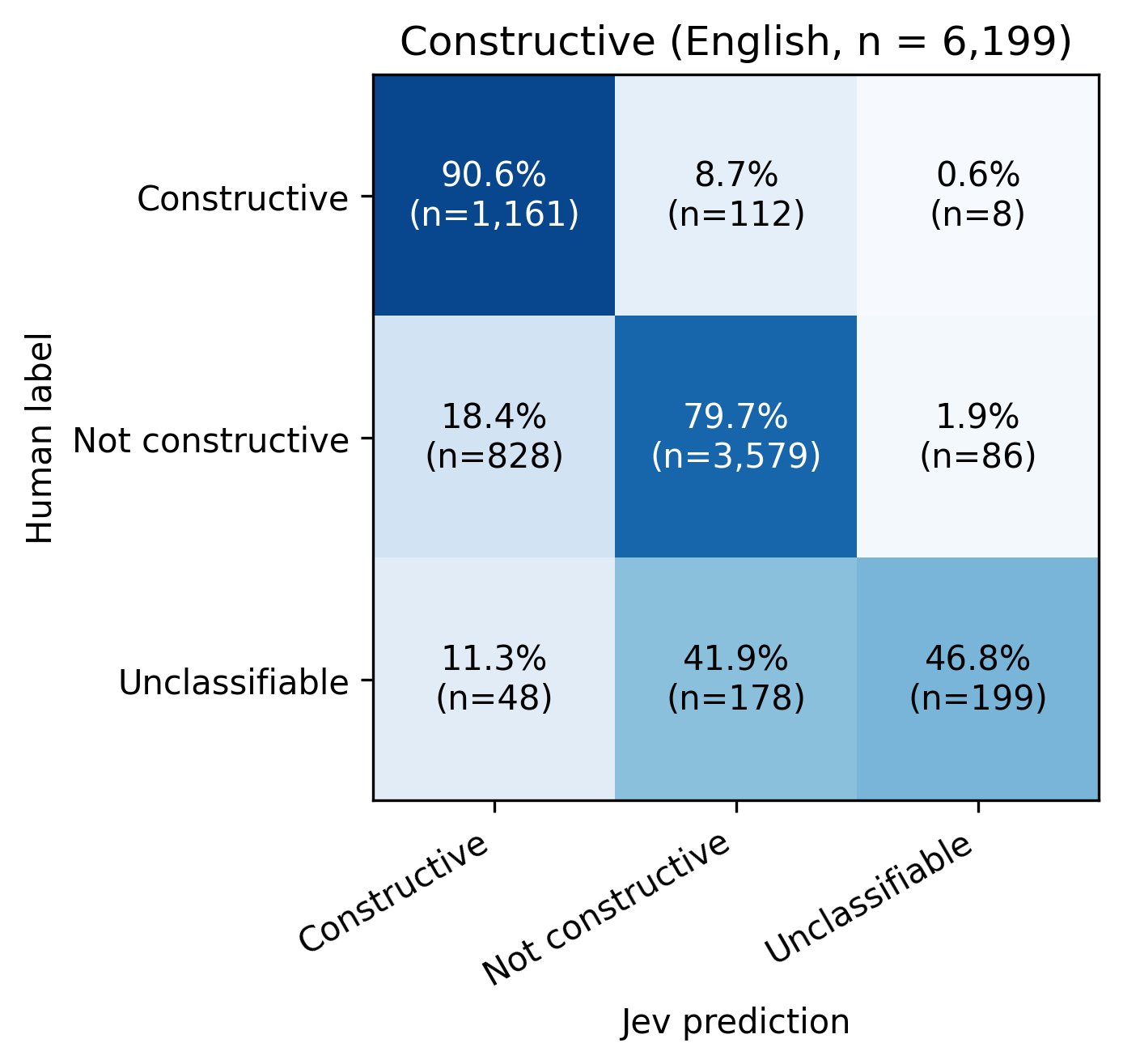}

    \vspace{0.2cm}

    \includegraphics[width=0.85\linewidth]{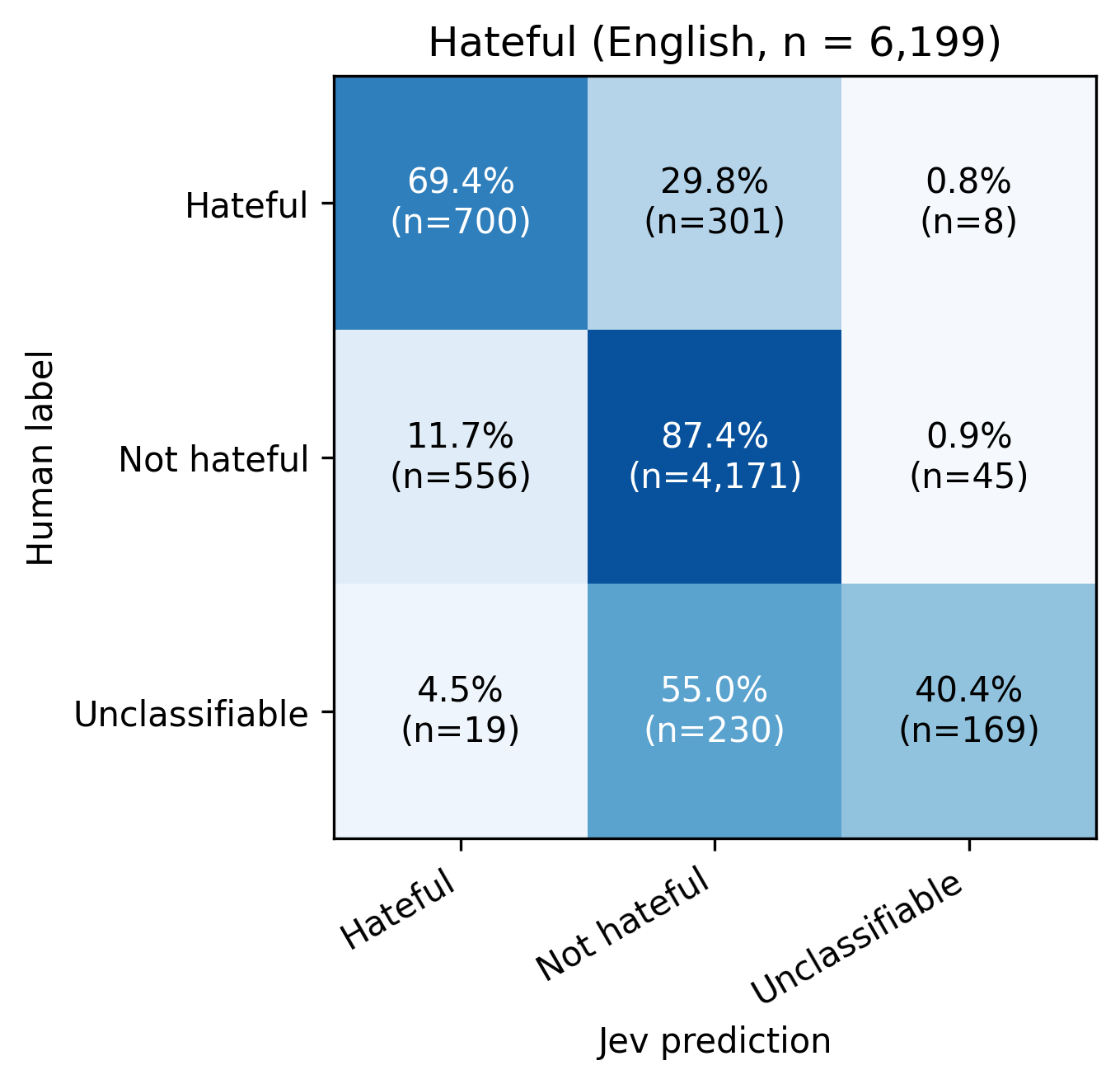}

    \caption{Confusion matrices for the Jev benchmark (English only): (a) agreement, (b) constructive speech, and (c) hateful speech.}
    \label{fig:xlm-confusion}
\end{figure}

\clearpage

\subsection{Danish performance}

\begin{table}[H]
\centering
\small
\setlength{\tabcolsep}{4pt}
\begin{tabular}{lrrrr}
\hline
\textbf{Class} & \textbf{P} & \textbf{R} & \textbf{F1} & \textbf{Support} \\
\hline
\multicolumn{5}{l}{\textbf{Constructive}} \\
Constructive & 0.228 & 0.681 & 0.342 & 160 \\
Not constructive & 0.874 & 0.638 & 0.738 & 972 \\
Unclassifiable & 0.000 & 0.000 & 0.000 & 57 \\
\textit{Macro avg} & \textit{0.368} & \textit{0.440} & \textit{0.360} & \textit{1,189} \\
\textit{Weighted avg} & \textit{0.746} & \textit{0.613} & \textit{0.649} & \textit{1,189} \\
\hline
\multicolumn{5}{l}{\textbf{Hateful}} \\
Hateful & 0.726 & 0.488 & 0.584 & 125 \\
Not hateful & 0.891 & 0.978 & 0.933 & 1,007 \\
Unclassifiable & 0.000$^{\dagger}$ & 0.000 & 0.000 & 57 \\
\textit{Macro avg} & \textit{0.539} & \textit{0.489} & \textit{0.505} & \textit{1,189} \\
\textit{Weighted avg} & \textit{0.831} & \textit{0.880} & \textit{0.851} & \textit{1,189} \\
\hline
\multicolumn{5}{l}{\textbf{Agreement}} \\
Agree & 0.497 & 0.863 & 0.631 & 307 \\
Disagree & 0.681 & 0.783 & 0.728 & 460 \\
Unclear & 0.811 & 0.244 & 0.375 & 422 \\
\textit{Macro avg} & \textit{0.663} & \textit{0.630} & \textit{0.578} & \textit{1,189} \\
\textit{Weighted avg} & \textit{0.680} & \textit{0.612} & \textit{0.578} & \textit{1,189} \\
\hline
\end{tabular}
\caption{Jev vs.\ human annotation, Danish ($n = 1{,}189$): precision (P), recall (R) and F1 per class; macro avg.\ is the unweighted mean over the three classes, weighted avg.\ is weighted by support. $^{\dagger}$\,Jev made no predictions in this class, so precision is undefined; it is set to 0 and included in the macro average.}
\label{tab:jev-da}
\end{table}

\begin{figure}[H]
    \centering

    \includegraphics[width=0.85\linewidth]{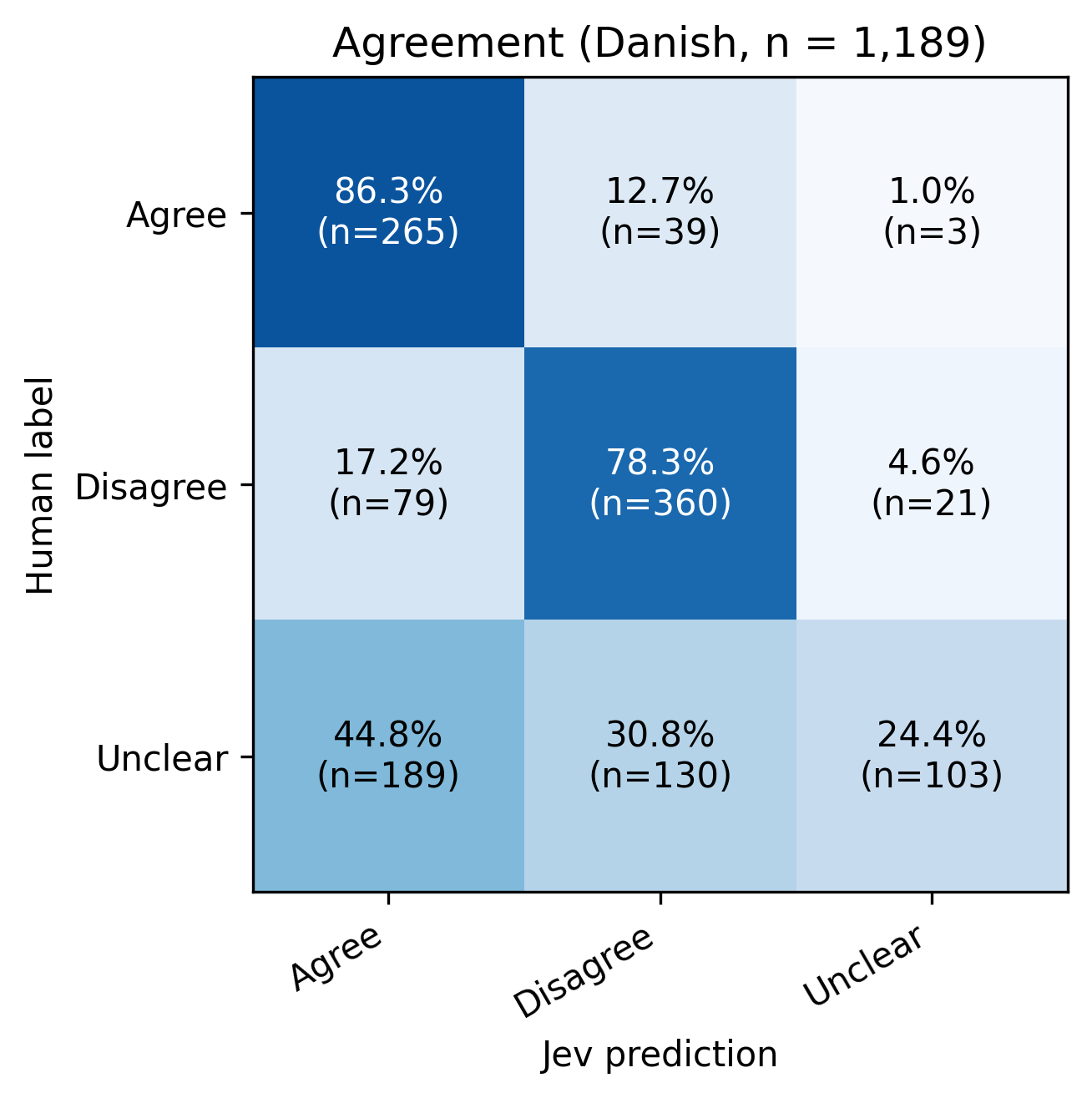}

    \vspace{0.2cm}

    \includegraphics[width=0.85\linewidth]{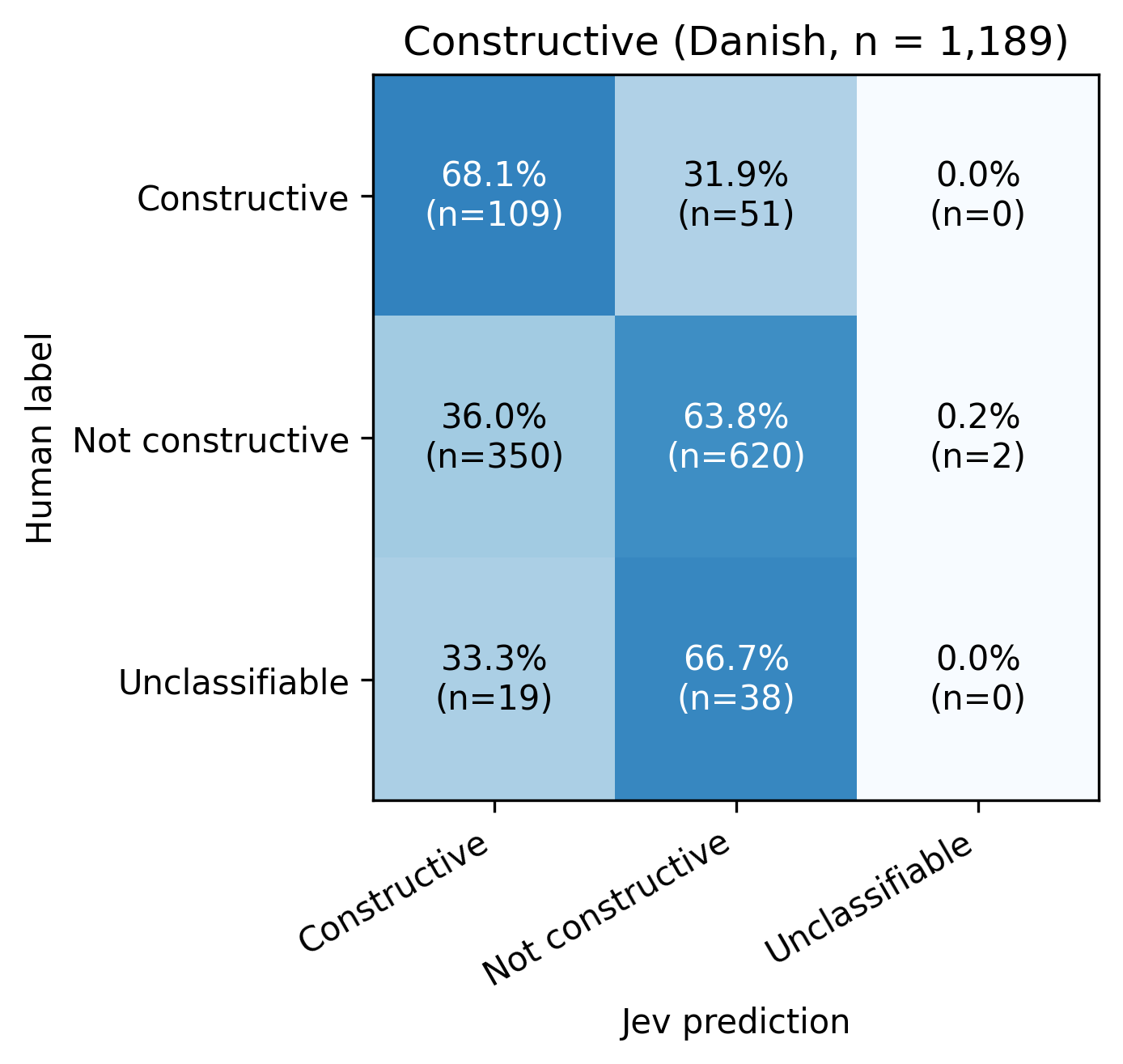}

    \vspace{0.2cm}

    \includegraphics[width=0.85\linewidth]{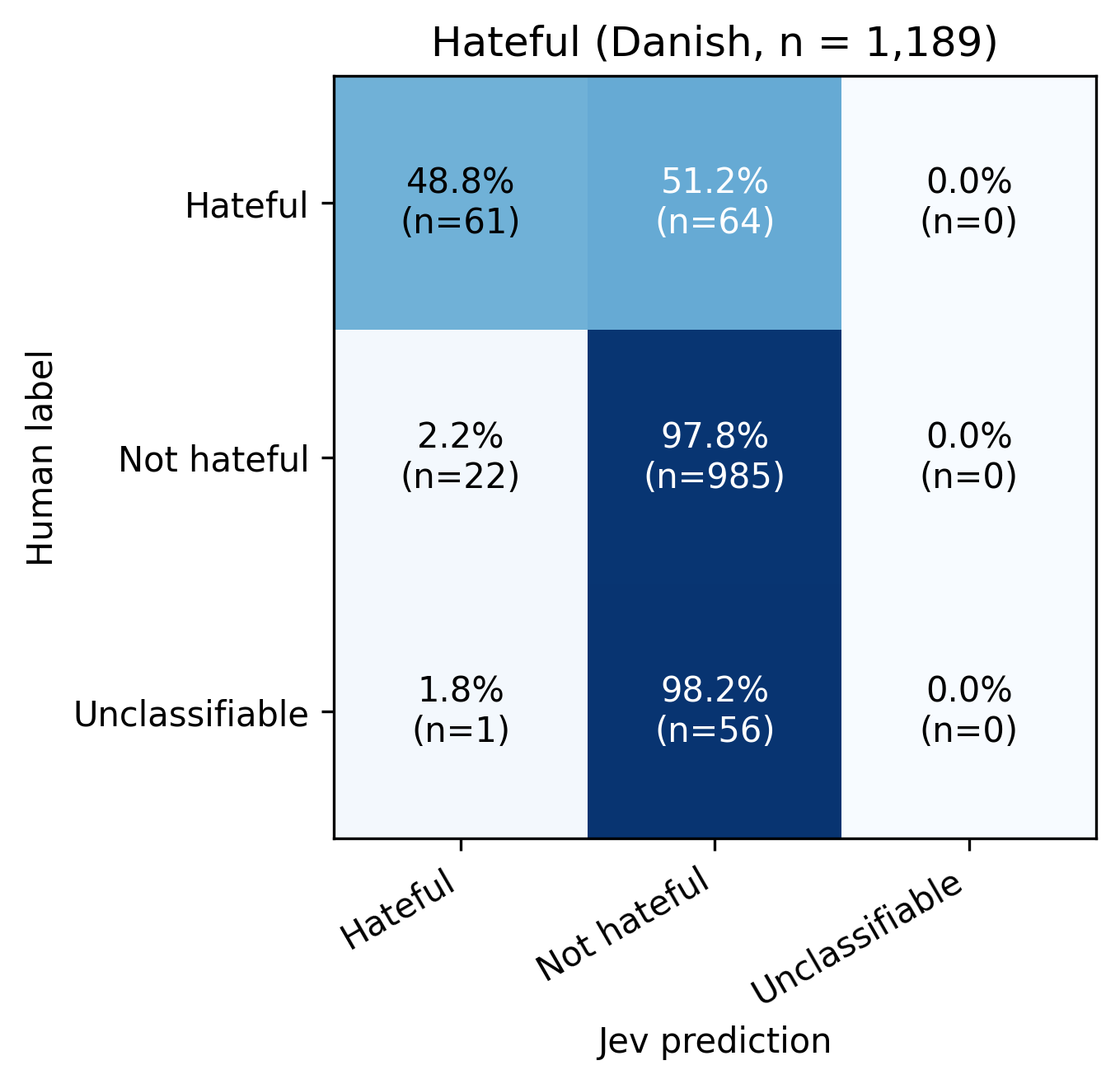}

    \caption{Confusion matrices for the Jev benchmark (Danish only): (a) agreement, (b) constructive speech, and (c) hateful speech.}
    \label{fig:xlm-confusion}
\end{figure}

\clearpage
\flushbottom
\end{document}